\documentclass[3p]{elsarticle}

\usepackage{amsmath,amssymb,amsfonts,amsthm}
\usepackage{multirow,algorithm,booktabs,color,footnote}
\usepackage[none]{hyphenat}
\usepackage{graphics,graphicx}
\usepackage{bm}
\usepackage{cases}
\usepackage{algorithmicx,algpseudocode}
\usepackage{xfrac}
\usepackage{rotating}

\usepackage{lscape}
\usepackage{float}

\usepackage[pagebackref=false,colorlinks]{hyperref}
\hypersetup{pdffitwindow=true,linkcolor=black,citecolor=black,urlcolor=black,menucolor=black}
\biboptions{sort&compress}

\newcommand{\op}[1]{\operatorname{#1}}

\newcommand{\Trans}[0]{^{\op{T}}}

\begin{document}

\begin{frontmatter}

\title{Tensor-Train Networks for Learning Predictive\\ Modeling of Multidimensional Data}

\author[DSPCom]{Michele Nazareth da Costa}
\ead{nazareth@decom.fee.unicamp.br}

\author[DSPCom]{Romis Attux}
\ead{attux@dca.fee.unicamp.br}

\author[cichocki1,cichocki2]{Andrzej Cichocki}
\ead{A.Cichocki@skoltech.ru}

\author[DSPCom]{João M. T. Romano}
\ead{romano@dmo.fee.unicamp.br}

\address[DSPCom]{DSPCom laboratory, University of Campinas, Campinas-SP, Brazil}

\address[cichocki1]{Skolkovo Institute of Science and Technology (SKOLTECH), CDISE, Moscow, Russian Federation}
\address[cichocki2]{Nicolaus Copernicus University, 87-100 Torun, Poland}

\begin{abstract}
In this work, we firstly apply the Train-Tensor (TT) networks to construct a compact representation of the classical Multilayer Perceptron, representing a reduction of up to 95\% of the coefficients. A comparative analysis between tensor model and standard multilayer neural networks is also carried out in the context of prediction of the Mackey-Glass noisy chaotic time series and NASDAQ index. We show that the weights of a multidimensional regression model can be learned by means of TT network and the optimization of TT weights is a more robust to the impact of coefficient initialization and hyper-parameter setting. Furthermore, an efficient algorithm based on alternating least squares has been proposed for approximating the weights in TT-format with a reduction of computational calculus, providing a much faster convergence than the well-known adaptive learning-method algorithms, widely applied for optimizing neural networks.
\end{abstract}


\end{frontmatter}

\section{Introduction}\label{sec:intro}
Deep neural networks have attracted the attention of the machine learning community because of their appealing data-driven framework and of their performance in several pattern recognition tasks. On the other hand, there are many open theoretical problems regarding the internal operation of the network, the necessity of certain layers, hyper-parameter selection etc. A promising strategy is based on tensor networks, which have been very successful in physical and chemical applications. In general, higher-order tensors are decomposed into sparsely interconnected lower-order tensors. This is a numerically reliable way to avoid the curse of dimensionality and to provide highly compressed representation of a data tensor, besides the good numerical properties that allow to control the desired accuracy of approximation.

The problem of time series analysis has been approached with the aid of strategies like stochastic models \cite{BoJeReLj:15,Pa:83}, artificial neural networks (ANNs) \cite{ZhPaHu:98,AdCo:98,Zh:12}, fuzzy systems and support vector machines (SVMs) \cite{CoVa:95,GeSuBaLaLaVaMoVa:01,CaTa:03,RaLuSa:03}, to name a few. An emblematic linear solution is based on the \emph{auto-regressive moving average} (ARMA) model, which combines the concept of \emph{auto-regressive} (AR) and \emph{moving-average} (MA) models. Nonlinear extensions of this type of solution are, for instance, the non-linear moving average model \cite{Ro:77} and the class of auto-regressive conditional heteroskedasticity (ARCH) models \cite{En:82}. 

A more general approach is to use universal approximators, like artificial neural networks. ANNs have been consistently employed in time series analysis, since the 1980s, in a plethora of practical scenarios \cite{ZhPaHu:98,AdCo:98,Zh:12,LiZo:20}. In the last decade, interest in these networks has dramatically increased due to the progress made in deep learning. This is certainly a consequence of the remarkable performance deep neural networks (DNNs) have reached in a variety of complex tasks, like pattern recognition, natural language processing, audio signal processing and planning / game playing.

DNNs are known to demand a vast amount of data to take full advantage of their multiple feature extracting layers, and typically have a number of parameters of the order of millions. To overcome the limitations inherent to modern DNNs, there is a need for the development of new architectures and associated fast learning algorithms and the application of special data formats for storing the parameters of such network. Current advances in NNs in most cases are associated with heuristic construction of the network architecture and applicable only to a particular problem. On the other hand, there is no understanding of the internal \emph{modus operandi} of the network, of the necessity or redundancy of certain layers, of the optimal methods to choose hyper-parameters, among others. A very promising approach is based on tensor networks (TNs) \cite{Ci:14,CiLeOsPhZhMa:16,CiLeOsPhZhSuMa:17,YaHo:17,EfHiLe:19,DeLiHaShXi:20,SuPeLiRaSu:20,GuDr:21}.

TNs are one of the most successful tools in quantum information theory, and are an efficient way of representing large volume of multi-dimensional data with an exponentially reduced number of parameters while maintaining accuracy of the approximation within many applications of interest \cite{Or:14}. TNs generally decompose higher-order tensors into sparsely interconnected matrices or lower-order tensors \cite{Ci:14}, through certain pattern of contractions. There are several methods based on TNs, providing a range of applicability, such as Matrix Product State (MPS), Tree Tensor networks, Projected Entangled Pair States (PEPS), Multi-scale Entanglement Renormalization Ansatz (MERA) tensor networks \cite{Or:14}. In the present work, we focus on one of the simplest tensor networks, the Tensor-Train network (TTN)\footnote{In Quantum Physics, it corresponds to an Matrix Product State (MPS) representation with open boundary conditions \cite{Or:14}.}, introduced by Oseledets and Tyrtyshnikov \cite{OsTy:09,Os:11}, which provides a very good numerical properties and the ability to control the approximation error by means the TT-rank.

Tensor networks have also been already used to compress weights of neural networks \cite{NoPoOsVe:15,YuZhAnYu:19,HaGrStSeGr:19,MuRaLiYaNi:20,KoLiKoKhFuAn:20}. In \cite{NoPoOsVe:15} the authors investigated perspectives of application of the TT architecture for compressing the weights matrix of fully connected layer of DNN, trained for classification tasks, and obtained a compression capacity of more than 200.000 times. In \cite{YuZhAnYu:19}, the authors used the TT network to represent a novel recurrent architecture based on higher-order tensor for multivariate forecasting and demonstrated $5\!\sim \!12\%$ improvements for long-term prediction over general recurrent neural network (RNN) and long short-term memory (LSTM) architectures. Similarly in \cite{MuRaLiYaNi:20}, the authors used TT networks to effectively compress LSTM networks with some gain or very little loss of performance on natural language tasks. The authors in \cite{ReSt:20} performed a supervised learning to solve regression task using the TT model in order to reduce the feature space representing the input data. Therefore, the TT network has been shown a promising neural network compression tool, thanks its ability to compress while preserving the model performance. 

In this study, we apply the TT network to construct a compact representation of the classical multilayer perceptron (MLP). In contrast to the algorithm employed in \cite{NoPoOsVe:15}, which is based on the stochastic gradient descent method, we apply a direct and non-iterative approach to the estimation of each TT-core tensor as the conventional solution for a general regression model. Differently from \cite{ReSt:20}, we adopt the standard alternating least squares (ALS) algorithm  with a stabilization technique via QR decomposition (similar to \cite{HoRoSc:12}) by including a shrinkage regularization method. From our formulation derived for the optimization problem, we propose a reduction in the computational cost required in the optimization of each TT-core using previous calculations and in the calculus of the pseudo-inverse through the use of the Generalized Singular Value decomposition (GSVD) \cite{GoLo:13} and exploitation of the sparse structure of the regularization matrix. Furthermore, we also apply the TT architecture to directly solve regression problems on a range of synthetic environments and real-world time series data and compare it to the performance obtained with MLPs, which are the most widely used ANNs for regression analysis. In our work we consider the prediction of two different scenarios: noisy chaotic time series, by means of Mackey-Glass equation and a real financial time series, given by NASDAQ index.

This paper is organized as follows. We start by Section \ref{sec:notations} by introducing our notations, operations, and briefly the TT-tensor representation. Section \ref{sec:ttmodel} describes and discusses the learning model based on TT networks, by proposing a reduction of computational calculus and by deriving a regularization matrix factor. Section \ref{sec:optimization} analyses the optimization framework and discusses an alternative strategy to reduce the computational cost of pseudo-inverse calculus. Section \ref{sec:regression} discusses some general considerations regarding tensor and neural networks. In Section \ref{sec:results}, a comparative analysis is carried out in the context of neural network recovery and non-linear predictions of two time series. Finally, Section \ref{sec:conclusions} presents some conclusions.

\section{Notation and Preliminaries}\label{sec:notations}
The notation used here is similar to the one adopted in \cite{KoBa:09}. Scalars, column vectors (or first-order tensors), matrices (or second-order tensors), and higher-order tensors (tensors of order three or higher) are written with lower-case, boldface lower-case, boldface upper-case, and calligraphic letters, i.e. ($a$, $\mathbf{a}$, $\mathbf{A}$, $\mathcal{A}$), respectively. Let $\mathbb{R}^{I_1\times I_2\times \cdots \times I_N}$ denote the tensor space of real $I_1\times I_2\times \cdots \times I_N$-tensors, for any integer $N\ge1$. Analogously to \cite{GoLo:13}, we are identifying the vector space of real $I$-vectors, i.e. $\mathbb{R}^I$, with $\mathbb{R}^{I\times 1}$ and so the members of $\mathbb{R}^I$ are column vectors. In this way, we refer to row vectors through the transpose of vectors, i.e. $\mathbf{a}\Trans \in \mathbb{R}^{1\times I}$. Each element of an $N$-order tensor $\mathcal{A}\in \mathbb{R}^{I_1\times I_2\times \cdots \times I_N}$ is denoted by $\left[\,\mathcal{A}\,\right]_{i_1,i_2,\ldots,i_N}:=a_{i_1,i_2,\ldots,i_N}$, where $i_n\in \{1,\ldots,I_n\}$ with $n\in \{1,\ldots,N\}$. 

For a matrix $\mathbf{A}\in \mathbb{R}^{I_1\times I_2}$, we can denote the $k$-th  column and row respectively as $\mathbf{a}_{:\,k}\in \mathbb{R}^{I_1}$ for $k\in\{1,\ldots,I_2\}$ and $\mathbf{a}_{k\, :}\in \mathbb{R}^{I_2}$ for $k\in\{1,\ldots,I_1\}$. We denote as $\mathbf{A}_{:K_1,:K_2}\in \mathbb{R}^{K_1\times K_2}$ a sub-matrix of $\mathbf{A}\in \mathbb{R}^{I_1\times I_2}$ with row index varying from 1 to $K_1$ and column index varying from 1 to $K_2$, for $K_1\in\{1,\ldots,I_1\}$ and $K_2\in\{1,\ldots,I_2\}$. For a third-order tensor $\mathcal{A}\in \mathbb{R}^{I_1\times I_2\times I_3}$, we can denote the $k$-th slice of $\mathcal{A}$ by $\mathbf{A}_{k\, :\, :}\in \mathbb{R}^{I_2\times I_3}$ for $k\in\{1,\ldots,I_1\}$, $\mathbf{A}_{:\, k\, :}\in \mathbb{R}^{I_1\times I_3}$ for $k\in\{1,\ldots,I_2\}$, and $\mathbf{A}_{:\,:\, k}\in \mathbb{R}^{I_1\times I_2}$ for $k\in\{1,\ldots,I_3\}$ by fixing the $k$-th index of the first, second and third dimension, respectively. $\mathbf{A}\Trans$ and $\mathbf{A}^{-1}$ stand for transpose and inverse matrices of $\mathbf{A}$, respectively. $\mathbf{I}_N$ is the identity matrix of order $N$, $\mathcal{N}\!\left(\cdot\right)$ denotes a null-space of a matrix, $\left\|\cdot \right\|_2$ is the Euclidean norm, $\left\|\cdot \right\|_{\mathrm{F}}$ is the Frobenius norm.

The inner product (or scalar product) of two the same-sized tensors $\mathcal{A}, \mathcal{B}\in \mathbb{R}^{I_1\times I_2\times \cdots \times I_N}$, which can be seen as a direct extension of the classical inner product of two vectors, is defined as 
\begin{align*}
	\left<\mathcal{A},\mathcal{B}\right> &:= \sum\limits_{i_1=1}^{I_1} \sum\limits_{i_2=1}^{I_2} \cdots \sum\limits_{i_N=1}^{I_N} a_{i_1,i_2,\ldots,i_N} b_{i_1,i_2,\ldots,i_N}\\
	&= \sum\limits_{i_1=1}^{I_1} \sum\limits_{i_2=1}^{I_2} \cdots \sum\limits_{i_N=1}^{I_N} \left[\,\mathcal{A} * \mathcal{B}\,\right]_{i_1,i_2,\ldots,i_N},
\end{align*}
which can be rewritten in terms of the Hadamard product of two the same-sized tensors $\mathcal{A}$ and $\mathcal{B}$, denoted by $*$, also known as the element-wise product. 

The outer product is denoted by $\circ$ and the outer product of $N$ vectors is defined, element-wise, as 
\begin{align*}
	\left[\,\mathbf{a}^{(1)}\, \circ\, \mathbf{a}^{(2)}\, \circ\, \cdots\, \circ\, \mathbf{a}^{(N)}\,\right]_{i_1,i_2,\ldots,i_N} &:= a^{(1)}_{i_1} a^{(2)}_{i_2} \ldots a^{(N)}_{i_N},
\end{align*}
for all index values with $i_n\in\{1,\ldots,I_n\}$ and each $n$-th vector $\mathbf{a}^{(n)}\in \mathbb{R}^{I_n}$. Note that this product $\mathbf{a}^{(1)}\, \circ\, \mathbf{a}^{(2)}\, \circ\, \cdots\, \circ\, \mathbf{a}^{(N)}$ leads to an $N$-order rank-one tensor with size $I_1\times I_2\times \cdots \times I_N$.
  
The operator $\op{vec}\left(\cdot\right)$ forms a vector by stacking the modes of its argument (matrix or tensor), such that $\op{vec}\left(\mathbf{A}\right)\in \mathbb{R}^{I_1 I_2}$ for any matrix $\mathbf{A}\in \mathbb{R}^{I_1\times I_2}$ or $\op{vec}\left(\mathcal{A}\right)\in \mathbb{R}^{I_1 I_2\cdots I_N}$ for any tensor $\mathcal{A}\in \mathbb{R}^{I_1\times I_2\times \cdots \times I_N}$. By convention adopted in the present work, the order of dimensions in a product, e.g. $I_1 I_2\cdots I_N$, is essentially linked to the order of variation of the corresponding index $(i_1,i_2,\ldots,i_N)$, such that the indexes placed more to the left vary slower and the ones placed more to the right vary faster. It will be important and essential in deriving the expressions presented throughout this work. 

The Kronecker product of matrices $\mathbf{A}\in \mathbb{R}^{I_1\times I_2}$ and $\mathbf{B}\in \mathbb{R}^{J_1\times J_2}$ is defined as
\begin{align*}
    \mathbf{A}\otimes \mathbf{B}:=\left[\!\! {\begin{array}{*{20}c}
    a_{1,1}\mathbf{B} \!\!& \cdots &\!\! a_{1,I_2}\mathbf{B}\\
    \vdots & \ddots & \vdots \\
    a_{I_1,1}\mathbf{B} \!\!& \cdots &\!\! a_{I_1,I_2}\mathbf{B}\\
    \end{array}} \!\!\right]\in \mathbb{R}^{I_1 J_1\times I_2 J_2}.
\end{align*}
The Khatri-Rao product (also called a column-wise Kronecker product) of matrices $\mathbf{A}\in \mathbb{R}^{I\times K}$ and $\mathbf{B}\in \mathbb{R}^{J\times K}$ is denoted by $\mathbf{A}\diamond \mathbf{B}$ and can be written in terms of the Kronecker product according to  
\begin{align*}
	\mathbf{A}\diamond \mathbf{B} = \left[\!\! {\begin{array}{*{20}c}
    \mathbf{a}_{:\, 1}\otimes \mathbf{b}_{:\, 1} \!\!& \cdots &\!\! \mathbf{a}_{:\, K}\otimes \mathbf{b}_{:\, K} \end{array}} \!\!\right] \in \mathbb{R}^{IJ\times K},
\end{align*}
For any $\mathbf{A}\in \mathbb{R}^{I\times J}$, $\mathbf{B}\in \mathbb{R}^{J\times L}$, $\mathbf{C}\in \mathbb{R}^{L\times M}$, a useful Kronecker property is given by
\begin{gather}\label{eq:prop0}
    \op{vec}\!\left(\mathbf{A}\mathbf{B}\mathbf{C}\right)=\left(\mathbf{A} \otimes \mathbf{C}\Trans\right) \op{vec}\!\left(\mathbf{B}\right)\in \mathbb{R}^{IM}.
\end{gather}

The \emph{unfolding} or \emph{matricization}, denoted by $\op{unfold}_n\left(\mathcal{A}\right)$ or $\mathbf{A}_n$, is the process of reordering the elements of a higher-order tensor $\mathcal{A}\in \mathbb{R}^{I_1\times I_2\times \cdots \times I_N}$ into a matrix with size $I_n\times I_1\cdots I_{n-1} I_{n+1} \cdots I_N$, by isolating the $n$-th mode of $\mathcal{A}$ and concatenating the remaining modes for any $n\in\{1,\ldots,N\}$, so that each element is given by
\begin{gather*}
	\left[\,\op{unfold}_n\left(\mathcal{A}\right)\,\right]_{i_n, \overline{i_1 \cdots i_{n-1} i_{n+1} \cdots i_N}} := a_{i_1,\ldots,i_n,\ldots,i_N},
\end{gather*}
for all index values and regarding the following definition
\begin{equation}\label{eq:def_multi-index}
\begin{aligned}
	\overline{i_1 \cdots i_N} &:= \sum\limits_{k=2}^N (i_{k-1}-1) \prod\limits_{l=k}^N I_l + i_N\\ 
	&\; = (i_1-1) I_2 \cdots I_N + \cdots + (i_{N-2}-1) I_{N-1} I_N + (i_{N-1}-1) I_N + i_N.
\end{aligned}
\end{equation}
Remark the order of the indexes in the above definition determines the order of variation of the corresponding index. The reverse process of unfolding is given by the operator $\op{fold}_n\left(\mathbf{A}_n,I_1\times \cdots \times I_N\right)$, which forms a tensor $\mathcal{A}\in \mathbb{R}^{I_1\times \cdots \times I_N}$ by unstacking the modes of its input matrix argument $\mathbf{A}_n\in \mathbb{R}^{I_n\times I_1\cdots I_{n-1} I_{n+1} \cdots I_N}$ according to the adequate dimension.

The $n$-mode product of a tensor $\mathcal{A}\in \mathbb{R}^{I_1\times I_2\times \cdots \times I_N}$ with a vector $\mathbf{x}\in \mathbb{R}^{I_n}$, defined as $\mathcal{A}\times_n \mathbf{x}$ with size ${I_1\times \cdots \times I_{n-1} \times I_{n+1}\times \cdots \times I_N}$ for $n \in\{1,\ldots,N\}$, represents a contraction of the $n$-th dimension of $\mathcal{A}$ to a low-order tensor given by
\begin{gather*}
	\left[\,\mathcal{A}\times_n \mathbf{x}\,\right]_{i_1,\ldots,i_{n-1},i_{n+1},\ldots,i_N} := \sum\limits_{i_n=1}^{I_n} a_{i_1,\ldots,i_n,\ldots,i_N}\, x_{i_n},
\end{gather*}
for all index values with $i_n\in \{1,\ldots,I_n\}$ and can be rewritten as follows
\begin{gather*}
	\op{vec}\left(\mathcal{A}\times_n \mathbf{x}\right) = \mathbf{x}\Trans \op{unfold}_n\left(A\right)\,\in\, \mathbb{R}^{I_1\cdots I_{n-1}I_{n+1}\cdots I_N}.
\end{gather*}

The $n$-mode product of a tensor $\mathcal{A}\in \mathbb{R}^{I_1\times I_2\times \cdots \times I_N}$ with a matrix $\mathbf{X}\in \mathbb{R}^{J\times I_n}$ is defined as $\mathcal{A}\times_n \mathbf{X}$ with size ${I_1\times \cdots \times I_{n-1} \times J \times I_{n+1}\times \cdots \times I_N}$ for $n \in\{1,\ldots,N\}$, such that each element is given by
\begin{gather*}
	\left[\,\mathcal{A}\times_n \mathbf{X}\,\right]_{i_1,\ldots,i_{n-1},j,i_{n+1},\ldots,i_N} := \sum\limits_{i_n=1}^{I_n} a_{i_1,\ldots,i_n,\ldots,i_N} x_{j, i_n},
\end{gather*}
for all index values with $i_n\in \{1,\ldots,I_n\}$ and $j\in \{1,\ldots,J\}$. It represents a linear transformation, mapping $\mathbb{R}^{I_n}$ to $\mathbb{R}^J$, on the $n$-the dimension of $\mathcal{A}$, such that
\begin{gather*}
	\mathcal{A}\times_n \mathbf{X} = \op{fold}_n\left(\mathbf{X}\, \mathbf{A}_n, I_1\times \cdots \times I_{n-1} \times J \times I_{n+1}\times \cdots \times I_N\right).
\end{gather*}

The $n$-mode canonical matricization of a tensor $\mathcal{A}\in \mathbb{R}^{I_1\times I_2\times \cdots \times I_N}$ results in a matrix $\mathbf{A}_{<n>}$ with size $I_1 I_2\cdots I_n \times I_{n+1}\cdots I_N$ and, using the definition in \eqref{eq:def_multi-index}, each element is given by 
\begin{align*}
	\left[\,\mathbf{A}_{<n>}\,\right]_{\overline{i_1 i_2  \cdots i_n},\overline{i_{n+1} \cdots i_N}} := a_{i_1,\ldots,i_n,\ldots,i_N}.
\end{align*}
As special cases, we have 
\begin{equation}
\begin{aligned}
	\mathbf{A}_{<1>}&=\mathbf{A}_1=\op{unfold}_1\left(\mathcal{A}\right), & \mathbf{A}_{<N-1>}&=\mathbf{A}_N\Trans=\left(\op{unfold}_{N}\left(\mathcal{A}\right)\right)\Trans, & \mathbf{A}_{<N>}&=\op{vec}\left(\mathcal{A}\right).\\
	&\in \mathbb{R}^{I_1\times I_2\cdots I_N} & &\in \mathbb{R}^{I_1\cdots I_{N-1}\times I_N} & &\in \mathbb{R}^{I_1 I_2\cdots I_N}
\end{aligned}
\end{equation}

\begin{figure*}[!h]
	\centering
	\def\svgwidth{0.85\columnwidth}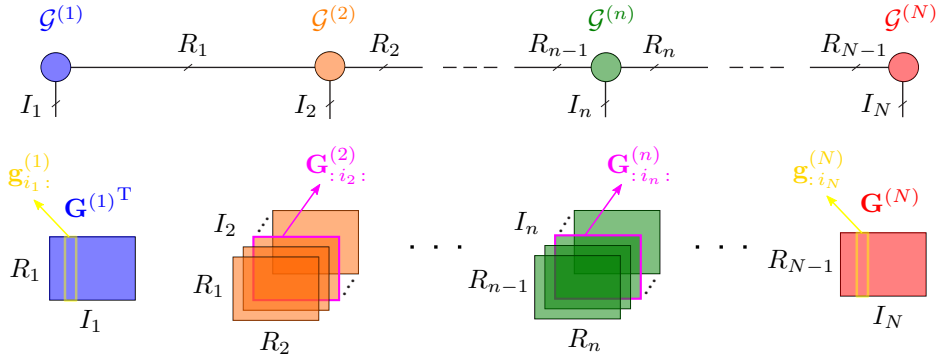
	\caption{Graphical representation of the Tensor-Train network for an $N$-th order data tensor. Since $R_0\!=\!R_N\!=\!1$, the core tensors $\mathcal{G}^{(1)}\in \mathbb{R}^{1\times I_1\times R_1}$ and $\mathcal{G}^{(N)}\in \mathbb{R}^{R_{N-1}\times I_N\times 1}$ can be directly rewritten as $\mathbf{G}_{1\,:\,:}^{(1)}:=\mathbf{G}^{(1)}\in \mathbb{R}^{I_1\times R_1}$ and $\mathbf{G}_{:\,:\, 1}^{(N)}:=\mathbf{G}^{(N)}\in \mathbb{R}^{R_{N-1}\times I_N}$, respectively. Both vectors $\mathbf{g}^{(1)}_{i_1\,:}\in \mathbb{R}^{R_1}$ and $\mathbf{g}^{(N)}_{:\,i_N}\in \mathbb{R}^{R_{N-1}}$ denote respectively the $i_1$-th row of $\mathbf{G}^{(1)}$ and the $i_N$-th column of $\mathbf{G}^{(N)}$.}\label{fig:TTnetwork}
\end{figure*}

In this study, we apply the Tensor-Train network \cite{Os:11} to represent a data tensor $\mathcal{X}\in \mathbb{R}^{I_1\times I_2\times \cdots \times I_N}$, as described in 
\begin{align}\label{eq:def_TTmodel_orig}
	x_{i_1,i_2,\ldots,i_N} &\cong \sum\limits_{r_1=1}^{R_1}\cdots \sum\limits_{r_{N-1}=1}^{R_{N-1}}\! g^{(1)}_{r_0,i_1,r_1} \cdots \,\,g^{(N)}_{r_{N-1},i_N,r_N}\, =\, {\mathbf{g}^{(1)}_{i_1\,:}}\Trans \mathbf{G}^{(2)}_{:\, i_2\,:} \cdots\: \mathbf{G}^{(N-1)}_{:\, i_{N-1} :}\; \mathbf{g}^{(N)}_{:\,i_N},   
\end{align}
where the tuple $\{R_1,\ldots,R_{N-1}\}$ is called the TT-rank and each tensor $\mathcal{G}^{(n)}\in \mathbb{R}^{R_{n-1}\times I_n\times R_n}$ denotes the TT-core for all $n\in\{1,...,N\}$ with $r_n\in \{1,\ldots,R_n\}$, $i_n\in \{1,\ldots,I_n\}$, and the boundary conditions given by $R_0\!=\!R_N\!=\!1$. This network can be graphically represented in Fig. \ref{fig:TTnetwork}.

\

The TT-rank is an important parameter of the TT network and determining the proper values for $\{R_1,\ldots,R_{N-1}\}$ is one of the main challenges in this network, having been studied in several papers
\cite{Os:11,PhCiUsTiLuMa:20,SeCiYoSh:20,SeCiPh:21}. The TT-rank determines memory requirements and allows to control the trade-off between representational power and computational complexity of the TT structure. According to \cite{Os:11,HoRoSc:12b}, a decomposition \eqref{eq:def_TTmodel_orig} for a given tensor $\mathcal{X}$ exists if the TT-rank satisfies $R_n\le \op{rank}\left(\mathbf{X}_{<n>}\right)$ and a quasi-optimal approximation, in terms of the Frobenius norm, in the TT-format for a given TT-rank can be obtained from the SVD\footnote{Singular Value Decomposition (SVD).}-based TT decomposition algorithm, introduced in \cite{Os:11}.

One successful class of methods to perform tensor approximation via the TT-format is based on a generalization of the well-known alternating least squares (ALS) algorithm. The idea behind ALS optimization (also known as one-site DMRG, DMRG1) \cite{Os:11,HoRoSc:12} is to proceed with global nonlinear optimization of the TT network through local linear optimizations, by updating only one core at a time while all other core tensors remain fixed. Alternatively, the modified ALS algorithm (referred to as two-site DMRG, DMRG2) \cite{HoRoSc:12} considers the optimization of a contraction of two consecutive core tensors (called \emph{super-core} or \emph{bond tensor}) at a time and subsequently estimates both tensors by a low-rank factorization. The main advantage of this modification is that the TT-ranks can be easily adapted to obtain a desired accuracy, despite being computationally more expensive \cite{HoRoSc:12,CiLeOsPhZhSuMa:17}. The monotonic convergence of ALS methods, under orthogonality constraints (introduced for practical reasons in \cite{Os:11}) to ensure the numerical stability of the method, is achieved through the gradual optimization of all core tensors along the network over several forward-backward sweeps, which has been discussed in \cite{RoUs:13}.

For a given data tensor $\mathcal{X}\in \mathbb{R}^{I_1\times I_2\times \cdots \times I_N}$, the number of coefficients to be stored by means the TT-format \eqref{eq:def_TTmodel_orig} increases linearly with the tensor order $N$ and $I:=\max\left(\{I_n\}_{n=1}^N\right)$, and quadratically in the maximum TT-rank bound $R:=\max\left(\{R_n\}_{n=1}^{N-1}\right)$, that is
\begin{align*}
	\sum\limits_{n=1}^N R_{n-1} I_n R_n \sim \mathcal{O}(N I R^2).
\end{align*}
In contrast to $\mathcal{O}(I^N)$ for the explicit storage of entries in $\mathcal{X}$, the memory requirements exponentially grows over the order of tensor data $N$ for a given $I$. Consequently, the TT network turns out an interesting alternative to overcome the curse of dimensionality. Another advantage of the TT structure is the simplicity of performing basic mathematical operations on tensors, directly considering $N$ tensors of order at most 3 (i.e., TT-cores $\{\mathcal{G}^{(n)}\}_{n=1}^N$) instead of an $N$-order dense tensor $\mathcal{X}$.

\section{Learning of Predictive Model}\label{sec:ttmodel}
In supervised machine learning, given a training dataset of pairs $\{\mathbf{x}^{(m)},y^{(m)}\}$, for $m\in \{1,\ldots,M\}$, where each input vector $\mathbf{x}^{(m)}$ is associated with a desired output $y^{(m)}$, the target output can be predicted according to the following model: 
\begin{align}\label{eq:def_decision_func1}
    {\hat y}^{(m)} := \left<\mathcal{W}, \,\mathit{\Phi}\!\left(\!\mathbf{x}^{(m)}\!\right)\right> = \sum\limits_{s_1=1}^{S_1}\cdots \sum\limits_{s_N=1}^{S_N} \left[\,\mathcal{W} * \mathit{\Phi}\!\left(\!\mathbf{x}^{(m)}\!\right)\,\right]_{s_1,\ldots,s_N},
\end{align}
where each $m$-th input vector $\mathbf{x}^{(m)}\! := \!\left[\!\!\begin{array}{ccc}
	x_1^{(m)}, \!\!& \!\!\ldots, \!\!& \!\!x_N^{(m)} \end{array}\!\!\right]\!\in \mathbb{R}^N$ is mapped onto a higher-order dimensional space through a feature map $\mathit{\Phi}\!: \mathbb{R}^N\! \to \mathbb{R}^{S_1\times\cdots \times S_N}$, and the tensor $\mathcal{W}\in \mathbb{R}^{S_1\times\cdots \times S_N}$ determines how each feature affects the prediction.
	
\begin{figure*}[!h]
	\centering
	\hspace{2cm}\def\svgwidth{0.85\columnwidth}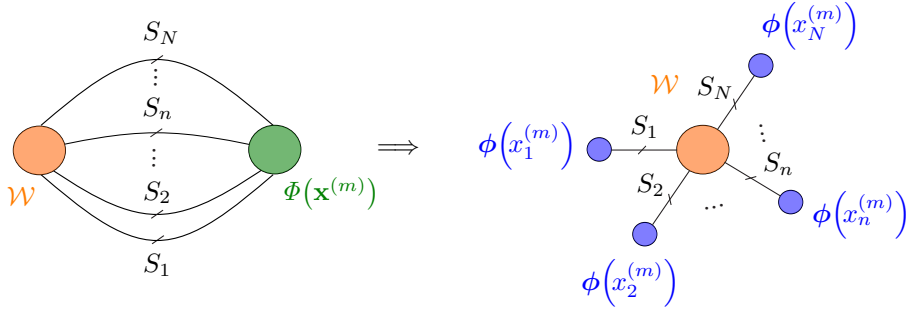
	\caption{Graphical illustration of the inner product in \eqref{eq:def_decision_func1} between two the same-sized tensors $\mathcal{W}$ and $\mathit{\Phi}\!\left(\mathbf{x}^{(m)}\right)$ (on the left side), and graphical representation of \eqref{eq:def_decision_func2} in terms of independent mappings associated to each $n$-th element $x_n^{(m)}$, denoted as $\boldsymbol{\phi}\!\left(\!x_n^{(m)}\right)\in \mathbb{R}^{S_n}$, for $n\in\{1,\ldots,N\}$ (on the right side).}\label{fig:TTregression}
\end{figure*}

We can simplify the previous model \eqref{eq:def_decision_func1} by considering independent mappings associated to each $n$-th element of the input vector $\mathbf{x}^{(m)}$, by $\boldsymbol{\phi}\!: \mathbb{R}\! \to \mathbb{R}^{S_n}$, as follows
\begin{equation}\label{eq:def_decision_func2}
\begin{aligned}
   {\hat y}^{(m)} &= \mathcal{W} \times_1 \boldsymbol{\phi}\!\left(\!x_1^{(m)}\right) \cdots \times_N \boldsymbol{\phi}\!\left(\!x_N^{(m)}\right)\\
   &= \sum\limits_{s_1=1}^{S_1}\cdots \sum\limits_{s_N=1}^{S_N} w_{s_1,\ldots,s_N}\, \left[\boldsymbol{\phi}\!\left(\!x_1^{(m)}\right)\right]_{s_1}\cdots \left[\boldsymbol{\phi}\!\left(\!x_N^{(m)}\right)\right]_{s_N},
\end{aligned}
\end{equation}
where $\boldsymbol{\phi}\!\left(\!x_n^{(m)}\right)\in \mathbb{R}^{S_n}$, for all $n\in\{1,\ldots,N\}$, and using the following relation
\begin{equation}\label{eq:def_decision_func_aux}
\begin{aligned}
    \left[\,\mathit{\Phi}\!\left(\!\mathbf{x}^{(m)}\!\right)\,\right]_{s_1,\ldots,s_N} &= \left[\boldsymbol{\phi}\!\left(\!x_1^{(m)}\right)\right]_{s_1}\cdots \,\left[\boldsymbol{\phi}\!\left(\!x_N^{(m)}\right)\right]_{s_N}\\
    \Longrightarrow\quad \mathit{\Phi}\!\left(\!\mathbf{x}^{(m)}\!\right) &= \boldsymbol{\phi}\!\left(\!x_1^{(m)}\right)\circ\, \cdots\, \circ\, \boldsymbol{\phi}\!\left(\!x_N^{(m)}\right),
\end{aligned}
\end{equation}
for all index values with $s_n\in\{1,\ldots,S_n\}$, which connects both expressions \eqref{eq:def_decision_func1} and \eqref{eq:def_decision_func2}. Observe that the model in \eqref{eq:def_decision_func2} is a particular case of \eqref{eq:def_decision_func1}, which is motivated by the encoding of local features. The selection of independent feature maps $\boldsymbol{\phi}$ will be discussed in more detail in Subsection \ref{ssec:encoding_data}. Fig. \ref{fig:TTregression} graphically represents both equations \eqref{eq:def_decision_func1} and \eqref{eq:def_decision_func2}. 	

\

Note that this model equation \eqref{eq:def_decision_func1} is linear with respect to the weight tensor $\mathcal{W}$ and can be seen as a straightforward extension of the classical linear regression model for higher-order dimensional data, to handle polynomials of any functions of input data, similarly to the one used in \cite{NoTrOs:16}. 
	
The most common method used for fitting regression problems is based on the least squares (LS) method \cite{Bi:06,Ha:13}. Thus, the predictors resulting from this model, i.e., those based on $\mathcal{W}$, can be learned by minimizing the mean squared error (MSE) function: 
\begin{align}\label{eq:def_cost_func}
	l\!\left(\mathcal{W}\right) = \frac{1}{M} \sum\limits_{m=1}^M\! \left(\left<\mathcal{W}, \,\mathit{\Phi}\!\left(\!\mathbf{x}^{(m)}\!\right)\right> - y^{(m)}\right)^2 = \frac{1}{M} \left\|\mathbf{\hat y} - \mathbf{y}\right\|^2_2,
\end{align}
where $\mathbf{y}\! := \!\left[\!\!\begin{array}{ccc}
	y^{(1)}, \!\!& \!\!\ldots, \!\!& \!\!y^{(M)} \end{array}\!\!\right]\! \in \mathbb{R}^M$ and $\mathbf{\hat y}\! := \!\left[\!\!\begin{array}{ccc}
	{\hat y}^{(1)}, \!\!& \!\!\ldots, \!\!& \!\!{\hat y}^{(M)} \end{array}\!\!\right]\! \in \mathbb{R}^M$
denote respectively the concatenation of all desired outputs and its predictions associated with the input vectors $\{\mathbf{x}^{(1)},\ldots,\mathbf{x}^{(M)}\}$.

Feature functions, as well as the weighting tensor, can be exponentially large. In our case, both $N$-th order tensors $\mathcal{W}$ and $\mathit{\Phi}$ have $S_1 S_2 \cdots S_N$ components. A simple way to reduce the number of coefficients of the tensor $\mathcal{W}$ is to represent it in the TT-format given in \eqref{eq:def_TTmodel_orig},
\begin{align}\label{eq:def_TTmodel}   
    w_{s_1,s_2,\ldots,s_N} &= \!\sum\limits_{r_1,\cdots, r_N-1}\! g^{(1)}_{r_0,s_1,r_1} \cdots \,\,g^{(N)}_{r_{N-1},s_N,r_N},
   \end{align}
where each core tensor, called TT-core, is denoted by $\mathcal{G}^{(n)}\in \mathbb{R}^{R_{n-1}\times S_n\times R_n}$ for all $n\in\{1,...,N\}$ with $r_n\in \{1,\ldots,R_n\}$, $s_n\in \{1,\ldots,S_n\}$, and $R_0\!=\!R_N\!=\!1$. By adopting the TT-format for $\mathcal{W}$ in \eqref{eq:def_TTmodel}, the inner product complexity in \eqref{eq:def_decision_func1} will be $\mathcal{O}(N S R^2)$ for $R:=\max\left(\{R_n\}_{n=1}^{N-1}\right)$ and $S:=\max\left(\{S_n\}_{n=1}^N\right)$ instead of $\mathcal{O}\left(S^N\right)$ in the raw tensor format.

Regarding a TT-format for weighting tensor $\mathcal{W}$ in \eqref{eq:def_TTmodel}, we can rewrite the expression in \eqref{eq:def_decision_func2} by isolating the $k$-th core $\mathcal{G}^{(k)}$ in terms of Kronecker products as follows
\begin{align}\label{eq:decision_func1a}	
    {\hat y}^{(m)}&=\prod\limits_{n=1}^N \mathcal{G}^{(n)} \times_n \boldsymbol{\phi}\!\left(x_n^{(m)}\right)\nonumber\\
	&= \mathcal{G}^{(k)} \times_1 \mathbf{p}_{k-1}^-\!\left(\!\mathbf{x}^{(m)}\right) \times_2 \boldsymbol{\phi}\!\left(\!x_k^{(m)}\right) \times_3 \mathbf{p}_{k+1}^+\!\left(\!\mathbf{x}^{(m)}\right)\nonumber\\
	&= \left< \mathbf{p}_{k-1}^-\!\left(\!\mathbf{x}^{(m)}\right)\otimes \boldsymbol{\phi}\!\left(\!x_k^{(m)}\right)\otimes \mathbf{p}_{k+1}^+\!\left(\!\mathbf{x}^{(m)}\!\right), \op{vec}\!\left(\!\mathcal{G}^{(k)}\right)\right>,
\end{align}
where both vectors $\mathbf{p}_{k-1}^-\!\left(\mathbf{x}^{(m)}\right)$ and $\mathbf{p}_{k+1}^+\!\left(\mathbf{x}^{(m)}\right)$ represent respectively the contraction of the left and right sides of the TT structure, i.e.
\begin{align}\label{eq:decision_func1b}  
	\begin{cases}    		
        \mathbf{p}_{k-1}^-\!\left(\mathbf{x}^{(m)}\right) := \prod\limits_{n=1}^{k-1} \!\mathcal{G}^{(n)} \times_n \boldsymbol{\phi}\!\left(\!x_n^{(m)}\right) \in \mathbb{R}^{R_{k-1}}\\
        \mathbf{p}_{k+1}^+\!\left(\mathbf{x}^{(m)}\right) := \prod\limits_{n=k+1}^{N} \!\!\mathcal{G}^{(n)} \times_{n} \boldsymbol{\phi}\!\left(\!x_{n}^{(m)}\right) \in \mathbb{R}^{R_k},
    \end{cases}
\end{align}

\begin{figure*}[!h]
	\centering
	\hspace{10cm}\def\svgwidth{0.99\columnwidth}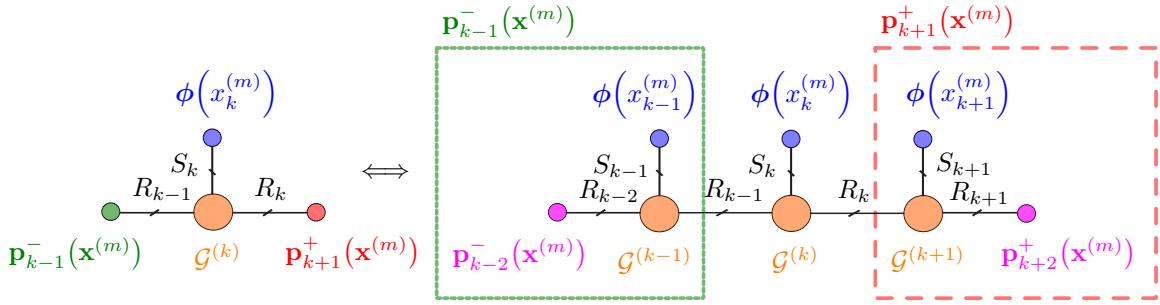
	\caption{Graphical illustration of the second equality in \eqref{eq:decision_func1a} (on the left side) and its equivalent representation by replacing both expressions in \eqref{eq:decision_func1c} into \eqref{eq:decision_func1a} (on the right side). Note that the contraction of the left and right sides of the TT-structure, i.e. for, respectively, all core tensors $\mathcal{G}^{(n)}$ with $n\in\{1,\ldots,k-2\}$ and $n\in\{k+2,\ldots,N\}$, is represented in magenta color.}\label{fig:TTregression_Kron3_opt}
\end{figure*}

Observe that both vectors $\mathbf{p}_{k-1}^-\!\left(\mathbf{x}^{(m)}\right)$ and $\mathbf{p}_{k+1}^+\!\left(\mathbf{x}^{(m)}\right)$ can be computed iteratively
\begin{align}\label{eq:decision_func1c}  
	\begin{cases} 
		\mathbf{p}_{k-1}^-\!\left(\mathbf{x}^{(m)}\right) = \mathcal{G}^{(k-1)}\times_1 \mathbf{p}_{k-2}^-\!\left(\mathbf{x}^{(m)}\right) \times_2 \boldsymbol{\phi}\!\left(\!x_{k-1}^{(m)}\right)\\
		\mathbf{p}_{k+1}^+\!\left(\mathbf{x}^{(m)}\right) = \mathcal{G}^{(k+1)}\times_2 \boldsymbol{\phi}\!\left(\!x_{k+1}^{(m)}\right)\times_3 \mathbf{p}_{k+2}^+\!\left(\mathbf{x}^{(m)}\right).
    \end{cases}
\end{align}
Thus by sweeping from left-to-right (or right-to-left), we can use $\mathbf{p}_{k-2}^-\!\left(\mathbf{x}^{(m)}\right)$ \big(or $\mathbf{p}_{k+2}^+\!\left(\mathbf{x}^{(m)}\right)$\big) to compute $\mathbf{p}_{k-1}^-\!\left(\mathbf{x}^{(m)}\right)$ \big(or $\mathbf{p}_{k+1}^+\!\left(\mathbf{x}^{(m)}\right)$\big) respectively. The use of \eqref{eq:decision_func1c} instead of \eqref{eq:decision_func1b} will reduce the demanding computational operations per each $k$-th core estimation, in terms of complex multiplications, once we can use the previous calculus of $\mathbf{p}_{k-2}^-\!\left(\mathbf{x}^{(m)}\right)$ \big(or $\mathbf{p}_{k+2}^+\!\left(\mathbf{x}^{(m)}\right)$\big). Hence, by computing $\mathbf{p}_{k-1}^-\!\left(\mathbf{x}^{(m)}\right)$ or $\mathbf{p}_{k+1}^+\!\left(\mathbf{x}^{(m)}\right)$ by means \eqref{eq:decision_func1c}, it leads to a complexity of $\mathcal{O}\left(SR^2\right)$. On the other hand, the calculus complexity of $\mathbf{p}_{k-1}^-\!\left(\mathbf{x}^{(m)}\right)$ and $\mathbf{p}_{k+1}^+\!\left(\mathbf{x}^{(m)}\right)$ by means \eqref{eq:decision_func1b} are respectively $\mathcal{O}\left((k-2)R^3 + (k-1)SR^2\right)$ and $\mathcal{O}\left((N-k-1)R^3 + (N-k)SR^2\right)$. This operation can be nicely represented in Fig. \ref{fig:TTregression_Kron3_opt}.

\

\renewcommand*{\thefootnote}{\fnsymbol{footnote}}

From the concatenation of all outputs $\{{\hat y}^{(1)},...,{\hat y}^{(M)}\}$, and by applying \eqref{eq:decision_func1a}, the estimated vector of the desired vector $\mathbf{y}$ can be expressed in terms of the $k$-th core $\mathcal{G}^{(k)}\in \mathbb{R}^{R_{k-1}\times S_k\times R_k}$, i.e. ${\boldsymbol \theta}_k$, by
\begin{align}\label{eq:output_estimate1a}
   \mathbf{\hat y} &= \left(\boldsymbol{\Phi}_{k}\diamond \mathbf{P}_{k-1}^-\diamond \mathbf{P}_{k+1}^+\right)\Trans \op{vec}\!\left(\!\mathbf{G}^{(k)}_2\!\right)\nonumber\\
   &= \mathbf{P}_k\, \boldsymbol{\theta}_k \in \mathbb{R}^M,
\end{align}
where $\mathbf{G}^{(k)}_2:=\op{unfold}_2\!\left(\mathcal{G}^{(k)}\right) \in \mathbb{R}^{S_k\times R_{k-1}R_k}$\footnote{Despite both vectors $\op{vec}\!\left(\!\mathbf{G}^{(k)}_2\!\right)\in \mathbb{R}^{S_k R_{k-1}R_k}$ and $\op{vec}\!\left(\mathcal{G}^{(k)}\!\right)\in \mathbb{R}^{R_{k-1} S_k R_k}$ contain the same elements of tensor $\mathcal{G}^{(k)}$, although in different positions, the preference in using the first vector instead of the second will be clear in the next section.} and
\begin{align}
    &\begin{cases}\label{eq:output_estimate1b}       
        &\boldsymbol{\Phi}_{k} := \left[\!\!\begin{array}{ccc}
        \boldsymbol{\phi}\!\left(x_k^{(1)}\right) \!\!& \!\!\cdots \!\!& \!\boldsymbol{\phi}\!\left(x_k^{(M)}\right)
        \end{array}\!\!\right] \in \mathbb{R}^{S_k\times M}\\
        &\mathbf{P}_{k-1}^- := \left[\!\!\begin{array}{ccc}
        \mathbf{p}_{k-1}^-\!\left(\mathbf{x}^{(1)}\right) \!\!& \!\!\cdots \!\!& \!\mathbf{p}_{k-1}^-\!\left(\mathbf{x}^{(M)}\right)
        \end{array}\!\!\right] \in \mathbb{R}^{R_{k-1}\times M}\\
        &\mathbf{P}_{k+1}^+ := \left[\!\!\begin{array}{ccc}
        \mathbf{p}_{k+1}^+\!\left(\mathbf{x}^{(1)}\right) \!\!& \!\!\cdots \!\!& \!\mathbf{p}_{k+1}^+\!\left(\mathbf{x}^{(M)}\right)
        \end{array}\!\!\right] \in \mathbb{R}^{R_k\times M}\\
		&\mathbf{P}_k := \left(\boldsymbol{\Phi}_{k}\diamond \mathbf{P}_{k-1}^-\diamond \mathbf{P}_{k+1}^+ \right)\Trans \in \mathbb{R}^{M\times S_k R_{k-1} R_k}\\
		&\boldsymbol{\theta}_k := \op{vec}\!\left(\!\mathbf{G}^{(k)}_2\right) \in \mathbb{R}^{S_k R_{k-1} R_k}.
    \end{cases}
\end{align}

Note that the remaining core tensors are absorbed by the matrix $\mathbf{P}_k$, from the above manipulations in accordance with \eqref{eq:decision_func1a}-\eqref{eq:decision_func1c}, and the $k$-the core tensor $\mathcal{G}^{(k)}$ is isolated in the expression \eqref{eq:output_estimate1a} with the aim of rewriting the loss function \eqref{eq:def_cost_func} in terms of the $k$-the core tensor. The importance of this procedure will become more clear in the next section.

Finally, the loss function, given in \eqref{eq:def_cost_func}, can be also expressed in terms of both vectors $\mathbf{y}$ and $\mathbf{\hat y}$, respectively associated with all target outputs and its predictions, by applying \eqref{eq:output_estimate1a} to \eqref{eq:def_cost_func} in the form
\begin{align}\label{eq:loss_regressor}
	l\left(\mathcal{W}\right) = \sfrac{1}{M} \left\|\mathbf{P}_k\, \boldsymbol{\theta}_k - \mathbf{y}\right\|^2_2.
\end{align}
	
If $\mathbf{P}_k$ has linearly independent columns, then $\mathbf{P}\Trans_k \mathbf{P}_k$ is non-singular matrix and the solution of least squares regression given by \eqref{eq:loss_regressor} turns out
\begin{align}\label{eq:solution_regressor}
	\boldsymbol{\hat \theta}_k &= \left(\mathbf{P}\Trans_k\, \mathbf{P}_k\right)^{-1} \mathbf{P}\Trans_k\,\mathbf{y}, 
\end{align}
where $\boldsymbol{\hat \theta}_k := \op{vec}\!\left(\!\mathbf{\hat G}^{(k)}_2\!\right)$ denotes an estimate of $\boldsymbol{\theta}_k$ and, consequently, an estimate of $\mathcal{G}^{(k)}$ since $\mathbf{\hat G}^{(k)}_2:=\op{unfold}_2\!\left(\mathcal{\hat G}^{(k)}\right)$.

\subsection{Shrinkage regularization method}
The collinearity (or multicolinearity) phenomenon affects calculations regarding individual predictors, in the sense that one predictor can be linearly determined through the others with a substantial degree of accuracy which leads to an inversion problem due to rank deficient of $\mathbf{P}_k$. In order to ensure that $\mathbf{P}\Trans_k \mathbf{P}_k$ is not ill-conditioned due to correlated columns of $\mathbf{P}_k$, i.e. collinear rows of $\mathbf{P}_{k-1}^-$, $\boldsymbol{\Phi}_{k}$, and $\mathbf{P}_{k+1}^+$ owing to Khatri-Rao structure given by \eqref{eq:output_estimate1b}, we can consider a regularization term $r\!\left(\mathcal{W}\right)$ added to the loss function \eqref{eq:loss_regressor}. 
Thus, we are minimizing the following function:
\begin{align}\label{eq:loss_regressor_withreg}
	l'\!\left(\mathcal{W}\right)&=l\left(\mathcal{W}\right) + \lambda\, r\!\left(\mathcal{W}\right),
\end{align}
where $\lambda\ge0$ denotes the regularization or shrinkage factor.

One common option, initially motivated to stabilize the solution \eqref{eq:solution_regressor}, is based on the $l_2$-norm of the weighting coefficients, also referred to as \emph{Tikhonov regularization} \cite{KeStOr:91}. 
In statistical literature, it is also known as \emph{ridge regression} \cite{HoKe:70} and the regularization term can be given by
\begin{align}\label{eq:ridge_regressor1}
	r\!\left(\mathcal{W}\right)=\left<\mathcal{W}, \mathcal{W}\right>=\left\|\mathcal{W}\right\|^2_{\mathrm{F}}.
\end{align}

In order to obtain an explicit regularization expression in terms of ${\boldsymbol \theta}_k$, we can rewrite the scalar product in \eqref{eq:ridge_regressor1} by isolating the $k$-th core $\mathcal{G}^{(k)}$ and contracting recursively the remaining cores on the left side $1\le n\le k-1$ and on the right side $k+1\le n\le N$, respectively denoted by $\mathbf{\tilde G}^{(k-1)^-}\in \mathbb{R}^{S_1\cdots S_{k-1}\times R_{k-1}}$ and $\mathbf{\tilde G}^{(k+1)^+}\in \mathbb{R}^{S_{k+1}\cdots S_N\times R_k}$, which are recursively obtained according to
\begin{equation}\label{eq:regularization_factor1a}
\begin{aligned}
	&\mathbf{\tilde G}^{(n)^-} :=
	\begin{cases}
		{\mathbf{G}^{(1)}_{1\,:\,:}} \in \mathbb{R}^{S_1\times R_1}, & n=1\\
		\left(\op{unfold}_3\!\left(\mathcal{G}^{(n)}\times_1 \mathbf{\tilde G}^{(n-1)^-}\right)\right)\Trans \in \mathbb{R}^{S_1\cdots S_n\times R_n}, & 2\le\! n\! \le\! k-1
	\end{cases}\\
	&\mathbf{\tilde G}^{(n)^+} :=
	\begin{cases}
		\left(\op{unfold}_1\!\left(\mathcal{G}^{(n)}\times_3 \mathbf{\tilde G}^{(n+1)^+}\right)\right)\Trans \in \mathbb{R}^{S_n\cdots S_N\times R_{n-1}}, & k+1\le\! n\! \le\! N-1\\
		\mathbf{G}_{:\,:\,1}^{{(N)}\Trans} \in \mathbb{R}^{S_N\times R_{N-1}}, & n=N.
	\end{cases}
\end{aligned}
\end{equation}

Finally, we can represent the weight tensor $\mathcal{W}$, defined in \eqref{eq:def_TTmodel}, in terms of $\mathbf{\tilde G}^{(k-1)^-}$ and $\mathbf{\tilde G}^{(k+1)^+}$, from \eqref{eq:regularization_factor1a}, by means its $k$-th matrix unfolding as follow
\begin{align}
	\op{unfold}_k\!\left(\mathcal{W}\right) &= {\mathbf{G}^{(k)}_2}\left(\!\mathbf{\tilde G}^{(k-1)^-}\!\otimes \mathbf{\tilde G}^{(k+1)^+}\right)\Trans\nonumber\\
	&= {\mathbf{G}^{(k)}_2} \mathbf{B}\Trans_k\, \in \mathbb{R}^{S_k\times \prod\limits_{n=1\atop n\ne k}^N\!\! S_n}\quad \text{ with},\label{eq:coefficients1a}\\
	\mathbf{B}_k &:= \mathbf{\tilde G}^{(k-1)^-}\!\otimes \mathbf{\tilde G}^{(k+1)^+}\in \mathbb{R}^{\prod\limits_{n=1\atop n\ne k}^N\!\! S_n\times R_{k-1} R_k}.\label{eq:coefficients1c}
\end{align}
Observe that the order of the dimensions is quite relevant because it denotes the speed at which each mode changes. 

The vectorization of a higher-order tensor can be derived from the vectorization of a matrix unfolding of this tensor. By applying the Kronecker property \eqref{eq:prop0}, we can represent the above matrix \eqref{eq:coefficients1a} as a vector given by
\begin{align}\label{eq:coefficients1b}
	&\op{vec}\!\left(\op{unfold}_k\!\left(\mathcal{W}\right)\right) = \mathbf{L}_k \op{vec}\!\left(\!\mathbf{G}^{(k)}_2\right)=\mathbf{L}_k {\boldsymbol \theta}_k\,\, \in \mathbb{R}^{S_k\!\! \prod\limits_{n=1\atop n\ne k}^N\!\! S_n}, \text{ with}\nonumber\\
	&\mathbf{L}_k := \mathbf{I}_{S_k}\!\otimes\!\left(\mathbf{\tilde G}^{(k-1)^-}\!\otimes \mathbf{\tilde G}^{(k+1)^+}\right)\in \mathbb{R}^{S_k\!\! \prod\limits_{n=1\atop n\ne k}^N\!\! S_n\times S_k R_{k-1} R_k}\nonumber\\
	&\hspace{0.55cm}= \mathbf{I}_{S_k}\!\otimes\!\mathbf{B}_k.
\end{align}
From \eqref{eq:ridge_regressor1}-\eqref{eq:coefficients1b}, we can write the regularization term as a function of the $k$-th core $\mathcal{G}^{(k)}$, given by ${\boldsymbol \theta}_k$, according to 
\begin{align}\label{eq:ridge_regressor2}
	r\!\left(\mathcal{W}\right)&=\left\|\op{unfold}_k\!\left(\mathcal{W}\right)\right\|^2_{\mathrm{F}}=\left\|\op{vec}\!\left(\mathcal{W}\right)\right\|^2_2,\nonumber\\
	&=\left\|\mathbf{L}_k\, {\boldsymbol \theta}_k\right\|^2_2
\end{align}
and the gradient vector with respect to ${\boldsymbol \theta}_k$ is
\begin{align}\label{eq:gradient_regularization}
     \frac{\partial}{\partial \boldsymbol{\theta}_k} r\!\left(\mathcal{W}\right) &= 2\, \mathbf{L}\Trans_k\, \mathbf{L}_k\, {\boldsymbol \theta}_k = 2 \left(\mathbf{I}_{S_k} \!\otimes\! \mathbf{B}_k\Trans \mathbf{B}_k\right) {\boldsymbol \theta}_k.
\end{align}

Regarding the linear LS problem based on the loss function \eqref{eq:loss_regressor_withreg}, i.e. 
\begin{equation}\label{eq:prob_regression_regularization}
\begin{aligned}
	& \underset{{\boldsymbol \theta}_k}{\text{minimize}}
	& & \sfrac{1}{M} \left\|\mathbf{P}_k\, \boldsymbol{\theta}_k - \mathbf{y}\right\|^2_2 + \lambda \left\|\mathbf{L}_k\, {\boldsymbol \theta}_k\right\|^2_2
\end{aligned},
\end{equation}
and under the assumption that the null-spaces of $\mathbf{P}_k$ and $\mathbf{L}_k$ intersect only trivially, i.e. 
\begin{gather}\label{eq:condition_uniquesolution}
	\mathcal{N}\!\left(\mathbf{P}_k\right)\cap \mathcal{N}\!\left(\mathbf{L}_k\right)\!=\!\{\mathbf{0}\}\hspace{0.15cm} \Longleftrightarrow\hspace{0.15cm} \mathrm{rank}\left(\left[\!\!\begin{array}{c} \mathbf{P}_k\\ \mathbf{L}_k \end{array}\!\!\right]\right)\!=S_k R_{k-1} R_k,
\end{gather}
the LS problem \eqref{eq:prob_regression_regularization} has the unique solution for any $\lambda\!>\!0$ given by \cite{Lo:76,El:77,Ha:89}
\begin{subequations}
\begin{gather}
     \frac{\partial}{\partial \boldsymbol{\theta}_k}\, l'\!\left(\mathcal{W}\right) = \left(\sfrac{2}{M}\,\mathbf{P}\Trans_k\, \mathbf{P}_k + 2\lambda\,\mathbf{L}\Trans_k\, \mathbf{L}_k\right)\boldsymbol{\theta}_k - \sfrac{2}{M}\,\mathbf{P}\Trans_k\,\mathbf{y},\nonumber\\ 
     \left(\mathbf{P}\Trans_k\, \mathbf{P}_k + \lambda M\,\mathbf{L}\Trans_k\, \mathbf{L}_k\right)\boldsymbol{\theta}_k = \mathbf{P}\Trans_k\,\mathbf{y}\label{eq:solution_regressor_regularization_1a}\\
       \boldsymbol{\hat \theta}_k = \left(\mathbf{P}\Trans_k\, \mathbf{P}_k + \lambda M\,\mathbf{L}\Trans_k\, \mathbf{L}_k\right)^{-1} \mathbf{P}\Trans_k \,\mathbf{y}.\label{eq:solution_regressor_regularization_1b}
\end{gather}
\end{subequations}
In case the condition \eqref{eq:condition_uniquesolution} is not met, the solution \eqref{eq:solution_regressor_regularization_1b} is not unique.

For $\lambda\!>\!0$, it makes the problem non-singular, as the matrix we need to invert no longer has a determinant near zero in the sense that its eigenvalues are no longer near zero, which avoids imprecise estimation of the inverse matrix \cite{KeStOr:91}. Besides solving ill-posed optimization problems, the use of regularization, by adjusting $\lambda$, allows to control the model's capacity \cite{HaTiFr:09} in terms of robustness and flexibility, preventing \emph{under-fitting} and \emph{over-fitting} problems.

There are other common shrinkage methods, such as \emph{Lasso (Least Absolute Shrinkage and Selection Operator) regression} \cite{Ti:96}, which induces sparsity constraint, and \emph{Elastic net} \cite{ZoHa:05}, designed to overcome limitations of Lasso and preferred when several features are strongly correlated, besides several variants of Lasso penalty, developed to tackle certain optimization limitations and to address to particular problems \cite{HaTiFr:09}. Despite this variety of methods, our present work is restricted to ridge regression, since it tends to perform better than Elastic net in case the number of observations $M$ is greater than the number of model parameters $P$ \cite{ZoHa:05}.

\subsection{Feature map: Encoding input data}\label{ssec:encoding_data}
In machine learning, feature maps can be specified in accordance with certain learning tasks in order to exploit the correlation of information inherent into input data and better classify or estimate it. Thus, input data could implicitly encode a localization information with the purpose of associating set of pixels to detect more efficiently a particular object in an image for example. Furthermore, feature mapping can allow non-linearly separable data to become linearly separable by a hyper-plane in a higher-order dimension.

According to \eqref{eq:def_decision_func2}, the same local feature, defined by $\boldsymbol{\phi}\!: \mathbb{R} \to \mathbb{R}^{S_n}$, is applied to each input $x_n^{(m)}$. Fitting a linear regression model may not be adequate when interactions between variables are not inherently linear. However, the linear regression framework can still be used if the model is nonlinear but linear with respect to its parameters. This is possible by means of a transformation applied to each input, such as a power or logarithmic transformation for example. We can include logarithmic transformation of features by regarding exponential regression model. As an example, for a three-dimension array, $S_n=3$, we have
\begin{align}\label{eq:exponential_regr}
    \boldsymbol{\phi}\!\left(\!x_n^{(m)}\right) = \left[\!\!\begin{array}{ccccc}
        1 \!& \!x_n^{(m)}\!& \!\log\!\left(x_n^{(m)}\right)
        \end{array}\!\!\right] \in \mathbb{R}^{S_n}.
\end{align}

Another possible way of generating nonlinear interaction features is to consider a polynomial regression model of degree $S_n\!-1$, which can be expressed by the Vandermonde structure (for $S_n=3$) given by
\begin{align}\label{eq:polynonial_regr}
    \boldsymbol{\phi}\!\left(\!x_n^{(m)}\right) = \left[\!\!\begin{array}{ccccc}
1 \!& \!x_n^{(m)}\!& \!x_n^{{(m)}^2}\end{array}\!\!\right] \in \mathbb{R}^{S_n}.
\end{align}
Note that the first-order polynomial leads to a multiple linear model whereas higher order ($S_n\ge 3$) allows a better fit for polynomial curves.

Remark that, in our approach, each TT-core $\mathcal{G}^{(n)}\in \mathbb{R}^{R_{n-1}\times S_n\times R_n}$ is used for mapping the existing interactions between inputs per each categorical feature. Therefore, the number of cores is determined by the number of features for a given data and the feature map regards the structure of inputs by exploiting nonlinear relationships.

\section{Optimization Framework}\label{sec:optimization}
To design an efficient learning algorithm, the parameters of our model can be derived from minimizing the mean of squared residuals on the training set under the TT-rank constraint. From \eqref{eq:def_cost_func}, it leads to
\begin{equation}\label{eq:prob_general}
\begin{aligned}
	& \underset{\mathcal{W}}{\text{minimize}}
	& & \frac{1}{M} \sum\limits_{m=1}^M \left(\left<\mathcal{W}, \,\mathit{\Phi}\!\left(\!\mathbf{x}^{(m)}\!\right)\right> - y^{(m)}\right)^2\\
	& \text{subject to}
	& & \text{TT-rank}\!=\!R.
\end{aligned}
\end{equation} 

Since the TT-rank for the desired solution is unknown beforehand, this procedure relies on an initial guess for the TT-rank, then it can be updated during the optimization procedure. Two different procedures can be adopted for this update: First, to start with a maximum rank and then to gradually reduce it or, alternately, to start with a minimum rank and then gradually increase it according to a prescribed residual tolerance or threshold rank value \cite{CiLeOsPhZhSuMa:17}.

An alternative strategy is to convert the optimization problem \eqref{eq:prob_general} into independent linear least squares problems for adaptively estimating only one core tensor $\mathcal{G}^{(k)}$ at a time by sweeping along all core tensors from left-to-right and right-to-left, by fixing the remaining cores. According to the development made in Section \ref{sec:ttmodel}, we can rewrite the overall problem \eqref{eq:prob_general} with a regularization factor by using 
\eqref{eq:prob_regression_regularization} as the following optimization approach
\begin{equation}\label{eq:prob_global}
\begin{aligned}
	& \underset{{\boldsymbol \theta}_1,\ldots,{\boldsymbol \theta}_N}{\text{minimize}}
	& & \left\{\sum\limits_{k=1}^N \left\|\mathbf{P}_k\, \boldsymbol{\theta}_k - \mathbf{y}\right\|^2_2 + \lambda M
\left\|\mathbf{L}_k\, {\boldsymbol \theta}_k\right\|^2_2\right\}\\
	& \text{subject to}
	& & \text{TT-rank}\!=\!R.
\end{aligned}
\end{equation} 

To reduce the computational complexity effort required for evaluating the solution in \eqref{eq:solution_regressor_regularization_1b} for several values of $\lambda$, we can first apply the GSVD of the matrix pair $\left(\mathbf{P}_k,\mathbf{L}_k\right)$, proposed by Van Loan \cite{Lo:76} assuming $M \ge S_k R_{k-1} R_k$ and the condition in \eqref{eq:condition_uniquesolution}, which is given by
\begin{align}\label{eq:gsvd1}
	\begin{cases}
		\mathbf{P}_k = \mathbf{U}_P \mathbf{\Sigma}_P \mathbf{V}\Trans\\
		\mathbf{L}_k = \mathbf{U}_L \mathbf{\Sigma}_L \mathbf{V}\Trans,
	\end{cases}
\end{align} 
where $\mathbf{U}_P$ and $\mathbf{U}_L$ are orthogonal matrices, $\mathbf{\Sigma}_P$ and $\mathbf{\Sigma}_L$ are diagonal matrices and $\mathbf{V}$ is non-singular matrix. 

By replacing \eqref{eq:gsvd1} in \eqref{eq:solution_regressor_regularization_1a}, it leads to an equivalent minimization problem, after some manipulations regarding $\mathbf{z}_k := \mathbf{V}\Trans \boldsymbol{\theta}_k$, it gets
\begin{subequations}
\begin{gather}
	\left(\mathbf{\Sigma}\Trans_P\,\mathbf{\Sigma}_P + \lambda M\,\mathbf{\Sigma}\Trans_L\, \mathbf{\Sigma}_L\right)\mathbf{z}_k = \mathbf{\Sigma}\Trans_P\,\mathbf{U}_P\Trans\,\mathbf{y}\nonumber\\
    \mathbf{\hat z}_k = \left(\mathbf{\Sigma}\Trans_P\,\mathbf{\Sigma}_P + \lambda M\,\mathbf{\Sigma}\Trans_L\, \mathbf{\Sigma}_L\right)^{-1} \mathbf{\Sigma}\Trans_P\,\mathbf{U}_P\Trans \,\mathbf{y}\,,\label{eq:gsvd2a}\\
    \boldsymbol{\hat \theta}_k = {\mathbf{V}\Trans}^{-1}\, \mathbf{z}_k.\label{eq:gsvd2b}
\end{gather}
\end{subequations}
From \eqref{eq:gsvd2a}, the inverse calculation is reduced to the inverse of each element on the diagonal, the decomposition in \eqref{eq:gsvd1} and the inverse matrix in \eqref{eq:gsvd2b} are computed just once for several values of $\lambda$.

There are different approaches to compute the GSVD or based on the GSVD, such as those discussed in \cite{El:82,Ha:89,GoLo:13,MoReSg:07,DyRe:14}, with the aim of reducing the computational effort and/or exploiting the structure of the regularization matrix. In \cite{El:82,MoReSg:07}, the GSVD computations take advantage of the structure of the regularization matrix, in case it is a band matrix or an orthogonal projection operator respectively. Additionaly, Eldén in \cite{El:82} discussed an alternative way to solve \eqref{eq:prob_regression_regularization}, in case $\mathbf{L}_k$ is not square and invertible, by considering a weighted inverse matrix which allowed the transformation of the original problem to a standard-form problem. Unlike those cases, the paper \cite{DyRe:14} proposed, based on \cite{El:82}, a method for computation of the GSVD and the truncated GSVD (TGSVD), proposed by Hansen in \cite{Ha:89} which generalizes truncated SVD, when the regularization matrix does not have an exploitable structure. Furthermore, Dykes and Reichel presented in \cite{DyRe:14} an approach for reducing the matrix pair $\left(\mathbf{P}_k,\mathbf{L}_k\right)$ to a a pair of simpler matrices in order to reduce the GSVD computations.

Note that our regularization matrix $\mathbf{L}_k$, defined in \eqref{eq:coefficients1b}, is a Kronecker product between $\mathbf{I}_{S_k}$ and $\mathbf{B}_k$. Therefore, it is a band matrix that enables to exploit the sparseness of its structure in the numerical computation regarding the regularization matrix, in accordance with the approaches discussed in \cite{El:77,El:82,Bj:88}. This analysis was not included in the scope of our study once there are several works proposed on this topic, as commented below.

The algorithmic details of our proposed technique for multilinear regression model is presented in Algorithm \ref{alg:TTregressor}. Note that the estimation of each TT-core is conditioned by the knowledge of previous estimating cores and an intermediate orthogonalization step is included by the QR decomposition, applied to each reshaped TT-core tensor defined in step 8 (Algorithm \ref{alg:TTregressor}), with the aim of guaranteeing the left and right orthogonality property of TT cores and consequently, the algorithm stability \cite{SaOs:11,Os:11,RoUs:13}. The criteria for selecting $\lambda$ is detailed in the next sections.

Remark that each core estimation problem can be seen as a layer in the network model, from which inputs with information $\mathbf{x}^{(m)}, \forall m\in \{1,...,M\}$, flow forward through the network. Hence the estimation of each core propagates the initial information along all network taking into account one feature per layer and finally produces the desired output. During the training, the sweeping procedure, widely applied for approximating TT structure, also allows that the information flow backwards through the network. Thus it can be analogously associated with the back-propagation learning in artificial neural network.

\noindent\begin{minipage}{\textwidth}
\begin{algorithm}[H]\caption{TT-MR: Multilinear regression model}\label{alg:TTregressor}
\begin{algorithmic}[1]
\State \scalebox{0.9}[1] {Random initialize all cores $\{\mathcal{G}^{(1)},\ldots,\mathcal{G}^{(N)}\}$}
\State \scalebox{0.9}[1] {Compute encoded inputs $\{\mathbf{\Phi}_1,\ldots,\mathbf{\Phi}_N\}$ by using \eqref{eq:output_estimate1b}}
\While {stop condition is not met}
    \For { $k \text{ into the range } \{1,2,\ldots,N-1\}$}
		\State \scalebox{0.9}[1] {Compute $\mathbf{P}_{k-1}^-$ and $\mathbf{P}_{k+1}^+$ using \eqref{eq:decision_func1b} and \eqref{eq:output_estimate1b}}
		\State \scalebox{0.9}[1] {Select $\lambda$ according to the lowest cost function}
        \State \scalebox{0.9}[1] {Estimate $\mathcal{\hat G}^{(k)}$ from $\boldsymbol{\hat \theta}_k := \op{vec}\!\left(\!\mathbf{\hat G}^{(k)}_2\!\right)$ by solving \eqref{eq:solution_regressor_regularization_1a} or \eqref{eq:gsvd2b}}
        \State \scalebox{0.9}[1] {Compute QR decomposition from $\mathcal{\hat G}^{(k)}$\footnote{Remember that $\mathbf{\hat G}^{(k)}_n$, for $n\in\{1,2,3\}$, denotes the $n$-th matrix unfolding of $\mathcal{\hat G}^{(k)}$ obtained in step 7.}: ${\mathbf{\hat G}^{{(k)}\Trans}_3}\!\!\!=\!\mathbf{Q}\mathbf{R}$\footnote{\label{note_lefttoright}Sweeping update from left to right, such that, in step 4, $k\in \{1,2,\ldots,N-1\}$.} or $\mathbf{\hat G}^{{(k)}\Trans}_1\!\!\!=\!\mathbf{Q}\mathbf{R}$\footnote{\label{note_righttoleft}Sweeping update from right to left, such that, in step 4, $k\in \{N,N-1,\ldots,2\}$.}}
        \State \scalebox{0.9}[1] {Set $r=\min\!\left(\op{rank}\!\left(\mathbf{\hat G}^{(k)}_3\right),R\right)$\textsuperscript{\ref{note_lefttoright}} or  $r=\min\!\left(\op{rank}\!\left(\mathbf{\hat G}^{(k)}_1\right),R\right)$\textsuperscript{\ref{note_righttoleft}}}       
        \State \scalebox{0.9}[1] {Update $\mathcal{\hat G}^{(k)}$ from $\mathbf{Q}$\footnote{The matrix $\mathbf{Q}_{:,:r}$ is built taking into account the $r$ first columns of $\mathbf{Q}$.}, such that $\mathcal{\hat G}^{(k)}=\op{fold}_3\left(\mathbf{Q}_{:,:r}\Trans,R_{k-1}\times S_k\times r\right)$\textsuperscript{\ref{note_lefttoright}} or $\mathcal{\hat G}^{(k)}=\op{fold}_1\left(\mathbf{Q}_{:,:r}\Trans,r\times S_k\times R_k\right)$\textsuperscript{\ref{note_righttoleft}}}
	\If {$k=N-1$}
		\State \scalebox{0.9}[1] {Update the last sweeping core from $\mathbf{R}$\footnote{The matrix $\mathbf{R}_{:r,:}$ is built taking into account the $r$ first rows of $\mathbf{R}$.}, such that $\mathcal{\hat G}^{(k+1)}= \mathcal{\hat G}^{(k+1)}\times_1 \mathbf{R}_{:r,:}$\textsuperscript{\ref{note_lefttoright}} or $\mathcal{\hat G}^{(k-1)}= \mathcal{\hat G}^{(k-1)}\times_3 \mathbf{R}_{:r,:}$\textsuperscript{\ref{note_righttoleft}}}
	\EndIf 
    \EndFor
    \State \scalebox{0.9}[1] {Repeat the above loop in the reverse order}
\EndWhile
\State \Return \scalebox{0.9}[1] {$\mathcal{W}$ in TT-format with cores $\{\mathcal{G}^{(1)},\ldots,\mathcal{G}^{(N)}\}$}
\end{algorithmic}
\end{algorithm}
\renewcommand\footnoterule{}
\end{minipage}

\section{General Considerations}\label{sec:regression}
In regression analysis, it is quite usual to standardize the inputs before solving \eqref{eq:prob_global} i.e. reparametrization using centered inputs, in order to avoid multicollinearity issues, which could affect model convergence, and also meaningful interpretability of regression coefficients. Consequently, it leads to estimate coefficients of ridge regression model without intercept \cite{HaTiFr:09}.

The choice of adaptive learning-method algorithms is dependent on the optimization problem and the method robustness noticeably affects convergence. The focus of this work is mainly to compare tensor and neural networks in terms of their structures, by means of robustness, prediction performance and network complexity. Taking it into consideration, we limit our analysis to the standard Gradient Descent (GD) and to the Adaptive Momentum Estimation \cite{KiBa:14} (or \emph{Adam}) algorithms, because its popularity in the domain.

Differently from standard model parameters, \emph{hyper-parameters} are employed in most machine learning algorithms to control the behavior of learning algorithms and there is no a closed formula to uniquely determine them from data. In general, they are empirically set by searching for the best value by trial and error, such that regularization factor, dropout rate, parameters of optimization algorithm (e.g. learning rate, momentum term, decay rate), among others. A usual way to find the best hyper-parameters is to regard the validation set and a search interval; therefore, this procedure, properly described in Section \ref{sec:results}, is equivalently applied to both approaches.

In ANNs, non-linearity is commonly introduced by activation functions for modeling outputs of intermediate and/or final layers with the aim of computing more complex problems, which is valuable for most of ANN applications. This function is usually selected according to some heuristic rules or desired properties, our work is restricted to two common functions: rectified linear unit (shortly referred to as ReLU) and hyperbolic tangent (briefly referred to as \emph{Tanh}) functions. 

Analogous to the determination of the number of layers in neural networks, the optimal rank determination beforehand is a very challenging problem in TT networks, which has been studied in several papers \cite{Os:11,PhCiUsTiLuMa:20,SeCiYoSh:20,SeCiPh:21}. However, it is possible to adaptively or gradually increase the TT-rank in each iteration with the aim of obtaining a desired approximation accuracy \cite{HoRoSc:12,PhCiUsTiLuMa:20,SeCiYoSh:20,SeCiPh:21}. On the contrary, in our approach in Subsection \ref{subsec:recoveringMLP}, we set TT parameters (the TT-rank is constrained by $R$, i.e. $R\!=\!\max\left(R_1,\ldots,R_{N-1}\right)$, and the dimension array $S_n$ is fixed to $S$ for all $n$) in order to obtain a range of the number of coefficients and compare each approximation performance obtained for a fixed MLP. In Subsections \ref{subsec:Mackey-Glass} and \ref{subsec:NASDAQ}, the parameters are set in a way to compare both TT and MLP, by approximately fixing the same number of coefficients.

It is usual to evaluate the performance progression of neural networks in terms of epochs, such that every epoch considers the entire data set to update the neural network. In contrast, TT algorithms typically consider the convergence speed in terms of sweeps along all core tensors. In order to set a fair comparison between tensor and neural networks, we take into account the contribution of the entire data on the update of all weights and, in this sense, it is reasonable to put on the same level the algorithmic convergence according to epochs and sweeps.

\section{Simulation Results}\label{sec:results}
In order to evaluate and compare the performance of the models, we consider the MSE of predictions, which is given by the loss function, and three other common metrics employed on regression problems: the explained variance score (briefly referred to here as \emph{score}), which measures the discrepancy between target and its prediction in terms of the sample variance (i.e. the quality of the fit of a model on data), the sample Pearson correlation coefficient (shortly referred to as SPCC), which measures the linear correlation between both variables (target and its prediction) regarding the estimates of co-variances and variances, and the coefficient of determination (known as \emph{$R$-squared} or $R^2$), which measures the degree of linear correlation and it is unable to determine whether the predictions are biased. These metrics are given by the following expressions:
\begin{align*}\label{eq:def_metrics}
	\rho_{\mathrm{MSE}} &:= \frac{1}{M} \sum\limits^M_{m=1} \left(y^{(m)}_{\mathrm{target}} - {\hat y}^{(m)}\right)^2,\\
	\rho_{\mathrm{score}} &:= \frac{1 - \mathrm{var}\left(\mathbf{y}_{\mathrm{target}} - \mathbf{\hat y} \right)}{\mathrm{var}\left(\mathbf{y}_{\mathrm{target}}\right)},\\
	\rho_{\mathrm{SPCC}} &:= \frac{\sum\limits^M_{m=1} \left(y^{(m)}_{\mathrm{target}} - {\bar y}_{\mathrm{target}}\right) \left({\hat y}^{(m)} - {\bar{\hat y}}\right)}{\sqrt{\sum\limits^M_{m=1} \left(y^{(m)}_{\mathrm{target}} - {\bar y}_{\mathrm{target}}\right)^2} \sqrt{\sum\limits^M_{m=1} \left({\hat y}^{(m)} - {\bar{\hat y}}\right)^2}},\\
	\rho_{\mathrm{R^2}} &:= 1 - \frac{\sum\limits^M_{m=1} \left(y^{(m)}_{\mathrm{target}} - {\hat y}^{(m)}\right)^2}{\sum\limits^M_{m=1} \left(y^{(m)}_{\mathrm{target}} - {\bar y}_{\mathrm{target}}\right)^2},
\end{align*}
where $\mathrm{var}\!\left(\cdot\right)$ denotes the sample unbiased variance operator, and ${\bar y}_{\mathrm{target}}$ and ${\bar {\hat y}}_{\mathrm{target}}$ mean the sample mean of the vector of target $\mathbf{y}_{\mathrm{target}}$ and its prediction $\mathbf{\hat y}_{\mathrm{target}}$.

\subsection{Setting parameters}
The weights of tensor and neural networks are only learned from the training and validation sets and the inputs of both networks are scaled to fit the range $[-1,1]$. It is known that this scaling procedure can provide an improvement on the quality of the solution, as it ensures all inputs are treated equally in the regularization process and allows a meaningful range for the random starting weights \cite{HaTiFr:09}. The starting values for weights are usually chosen to be random values close to zero. A good practice is to initialize the weights following the uniform distribution in the range of $[-\delta, \delta]$, where $\delta\!\buildrel \Delta \over=\!\sfrac{1}{\sqrt{n}}$ and $n$ denotes the number of coefficients associated to each neuron, and the biases to be zero. In analogy, the coefficients of each core tensor are also initialized according to this practice, by regarding $n$ in terms of the number of coefficients of each $n$-th core tensor $\mathcal{G}^{(n)}$.

The stopping criterion is based on early stopping (in order to avoid over-fitting), which is defined as a minimum relative improvement of loss function, regarding the last two consecutive iterations and normalized by the previous value, until some tolerance is achieved. Thus, we impose a minimum relative improvement of $10^{-6}$ over, at least, $20\%$ of the maximum number of epochs or sweeps. In all simulations, the data is separated in three different sets for training ($60\%$), validation ($20\%$) and test ($20\%$). To validate and better understand different aspects regarding the neural and tensor networks, we consider three different experiments separately described in the following three subsections.

\subsection{Recovering multilayer perceptrons}\label{subsec:recoveringMLP}
Firstly, we consider a data set with 10000 samples generated by means of a neural network (10-200-1) with 10 inputs and 200 neurons in the hidden layer, totaling 2401 coefficients. The input matrix, randomly generated by a uniform distribution into the range [$-1,1$], is propagated in two layers: hidden and output layer. Both weights and biases of the neural network are drawn from a Gaussian distribution with zero-mean and standard deviation equal to 2. Two activation functions, ReLU and Tanh functions, are included in the intermediate layer. We consider a maximum number of sweeps equal to 12, since the algorithm convergence is achieved with less number of sweeps.

The regularization factor $\lambda$ is selected according to a searching step based on the known \emph{Golden-search section} (GSS) with a rough preliminary search regarding the given interval $\{2^n\!: n\!\in\! \mathbb{Z}, -10\!\leq\! n\! \leq\! 10\}$. Thus, the optimal regularization factor for each $k$-th core estimate is chosen by taking into account the lowest value of the loss function computed from the validation set.

The neural network output was recovered by the 10-th order TT decomposition by fixing a maximum TT-rank ($R$), considering several values, and two values of dimension array ($S\in \{2,3\}$, such that $S_n\!=\!S$ for $n\in \{1,\ldots,N\}$), regarding the local feature mapping $\mathbb{R}\to \mathbb{R}^{S}$ given by the polynomial regression in \eqref{eq:polynonial_regr}. Tables \ref{tab:recoverNN_tanh} and \ref{tab:recoverNN_relu}, for Tanh and ReLU functions respectively, show the average performance for all configurations, over 100 Monte Carlo simulations, in terms of MSE, \emph{score}, SPCC, and \emph{$R$-squared} at the convergence, for training, validation and test sets. 

According with Table \ref{tab:recoverNN_tanh}, we verify that the performance is improved with the increment of both model parameters $R$ and $S$ once more coefficients are employed. From 232 to 2728 coefficients, for $S\!=\!2$ with $R\!=\!4$ and $R\!=\!40$, we obtained an improvement over the test set of 4.92\% in terms of the explained variance score. Analogously for $S\!=\!3$ with $R\!=\!2$ and $R\!=\!12$, from 108 and 2556 coefficients, we got an improvement of 12.53\% over the test set. Note that the TT model for $S\!=\!3$ and $R\!=\!14$, with 3288 coefficients, does not provide a better \emph{score} than the one for $S\!=\!3$ and $R\!=\!12$, with 2556, thus more coefficients lead to a slight over-fitting of the model.

In contrast to the results for recovering the NN with Tanh function, Table \ref{tab:recoverNN_relu} shows a lower improvement with the increase of $R$ and $S$. From $R\!=\!20$ with $S\!=\!2$, i.e. from more than 1960 coefficients, the model does not offer a meaningful improvement over the test set, i.e. lower than four decimal places. From 232 to 1960 coefficients, for $S\!=\!2$ with $R\!=\!4$ and $R\!=\!20$, we have a gain over the test set of 1.24\% against 10.34\% for $S\!=\!3$ with $R\!=\!2$ and $R\!=\!12$ (implying the increase of 108 to 2556 coefficients). Analogously to Table \ref{tab:recoverNN_tanh}, we observe a soft trend of over-fitting from $R\!=\!12$ to $R\!=\!14$ with $S\!=\!3$, because more coefficients did not provide a better \emph{score} over the test set. 

\begin{figure*}[!h]
	\centering
	\includegraphics[width=.9\textwidth]{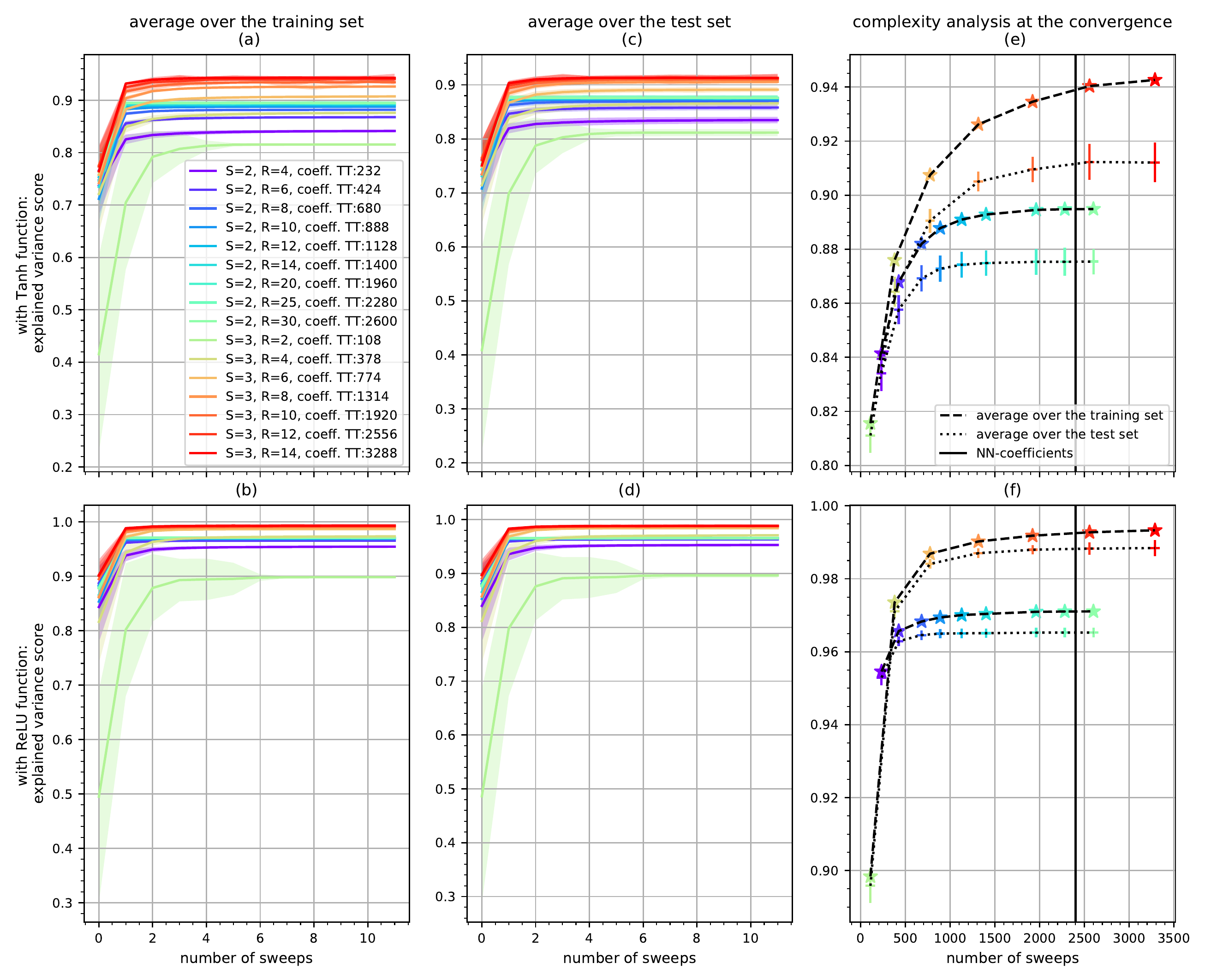}
	\caption{Recovery of (10-200-1) NN, with 10k samples (6k training + 2k validation + 2k test) and with Tanh \{(a),(c),(e)\} and ReLU \{(b),(d),(f)\} functions, using tensor-train network for different values of dimension array ($S$) and maximum TT-rank ($R$). Figs. \{(a),(b)\} and \{(c),(d)\} show the results for the training and test sets respectively. Figs. (e) and (f) show the complexity analysis at the convergence.}\label{fig:recoveryNN}
\end{figure*}

In Figure \ref{fig:recoveryNN}, we present the average \emph{score} over 100 Monte Carlo simulations, regarding all configurations, for the training and test sets. Note that the respective standard deviation is represented in this figure in order to stress the influence of the selection of sets and the initialization of the model coefficients. In accordance with Fig. \ref{fig:recoveryNN}, as previously discussed, more coefficients considered in the TT network lead to an improvement in the performance of the training set; in contrast with that, the performance of the test set tends to saturate from $R\!=\!14$ and $R\!=\!12$ for $S\!=\!2$ and $S\!=\!3$ respectively. In other words, the use of more than 1400 and 2556 coefficients for $S\!=\!2$ and $S\!=\!3$ does not improve the test set prediction - hence, to use more coefficients is pointless. 

It is interesting to observe the potential of contraction of the TT structures regarding a (10-200-1) NN with 2401 coefficients: it can be modeled as a TT network with much less coefficients. For $R\!=\!2$ and $S\!=\!3$, the TT network has only 108 coefficients, which represents less than 5\% of the total number of neural network coefficients, and can achieve an average \emph{score} for the test set equals to 0.8110 and 0.8958, regarding Tanh and ReLU functions. The best average performance for the test set is obtained for $S\!=\!3$ and $R\!=\!12$, with 2556 coefficients, with an average \emph{score} equal to 0.9126 and 0.9884 for, respectively, both Tanh and ReLU functions. 

Furthermore, Fig. \ref{fig:recoveryNN} also allows to better understand the influence of the parameter $S$, i.e. the dimension array of the encoded features. This parameter controls the degree level of the polynomial regression model, i.e. the level of non-linearity introduced by the feature mapping, and can enable to fit better the data interactions with lower number of coefficients, as shown in Fig. \ref{fig:recoveryNN}.

\subsection{Mackey-Glass noisy chaotic time series}\label{subsec:Mackey-Glass}
The Mackey–Glass system has been introduced as a model of white blood cell production \cite{MaGl:77}, which is usually modeled by delay-differential equations and provides a range of periodic and chaotic dynamics. Due to the dynamic properties and the mathematical simplicity, the Mackey-Glass time series has been employed to validate prediction methods, through the forecast of chaotic time series \cite{ChChMu:96,LeLoWa:01,GuWa:07,Mi:09,KoFuLiLe:11}. 

In the second experiment, we consider the Mackey-Glass noisy chaotic time series in order to compare both neural and tensor networks, which refers to the following delayed differential equation \cite{MaGl:77}:
\begin{equation}
	\frac{\delta x(t)}{\delta t} = x(t + \Delta t) = a\, \frac{x(t-\tau)}{1 + x(t-\tau)^n} - b\, x(t).
\end{equation}

The Mackey-Glass time series with 1000 samples was generated using the 4-th order Runge-Kutta method with the power factor $n\!\!=\!\!10$, initial condition $x(0)\!=\!1.2$, delay constant $\tau\!=\!17$, time step size $\Delta t\!=\!1.0$, and other parameters $a\!=\!0.2$ and $b\!=\!0.1$. According to \cite{Mi:09,Fa:82}, for $\tau\!\geq \!17$, the time series shows chaotic behavior. We consider four non-consecutive points of the time series, spaced by 6 points, with the aim of generating each input vector to predict the short-term $x(t+6)$ and long-term $x(t+84)$ predictions, i.e.
\begin{align*}
	x(t+6)&= F\!\left(x(t-18), x(t-12), x(t-6), x(t)\right)\\
	x(t+84)&= F\!\left(x(t-18), x(t-12), x(t-6), x(t)\right),
\end{align*}
which represents a usual test \cite{GuWa:07,Mi:09,KoFuLiLe:11}.
The noiseless case is considered, as well as experiments with additive white Gaussian noise with zero mean and two values of standard deviation i.e. $\sigma_{\mathrm{N}} \in \{0.05, 0.1\}$.

Three different 4-th order TT networks with ($S\!\!=\!\!2$,$R\!\!=\!\!2$), ($S\!\!=\!\!2$,$R\!\!=\!\!4$), ($S\!\!=\!\!3$,$R\!\!=\!\!4$) are employed to predict the short and long-term indices, as well as three different neural networks: (4-4-1), (4-6-1), (4-15-1) with two activation functions: Tanh and ReLU. The choice of these neural network parameters is due to the restriction of one hidden layer, as discussed above, and the TT parameters come from the approximate number of coefficients, i.e. (24, 40, 90) and (25, 37, 91) for the TT and NN structures respectively.

Analogously to the previous subsection, the regularization factor search for the tensor network follows the same described procedure, regarding the validation set i.e. it is based on the GSS with a rough preliminary search from the same interval. We also adopted this procedure for the neural networks in order to search an optimal learning rate applied on the SGD method.

\begin{figure*}[htb]
	\centering
	\includegraphics[width=.9\textwidth]{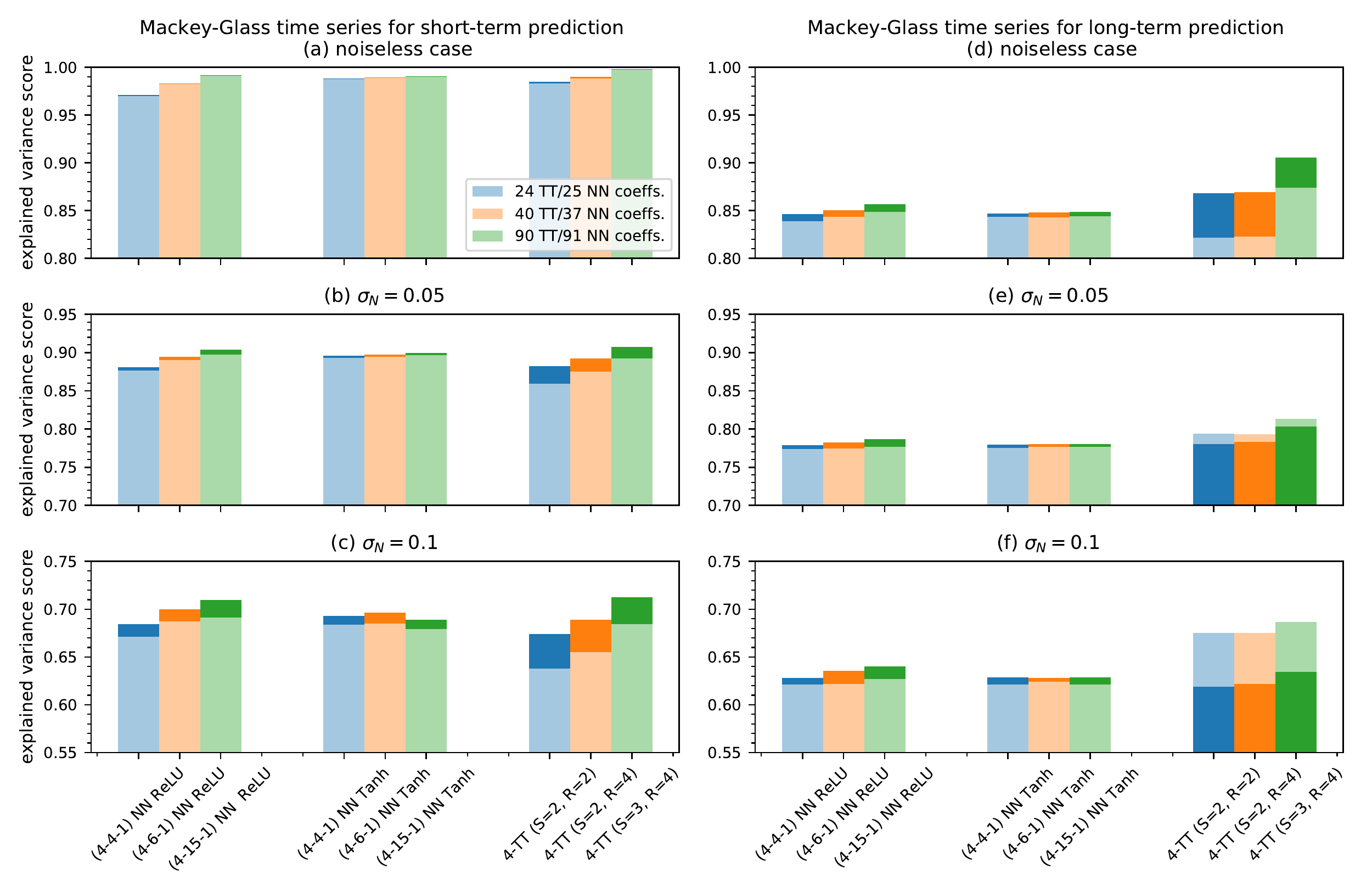}
	\caption{Mackey-Glass time series prediction regarding the noiseless case and the experiments with additive white Gaussian noise with zero mean and standard deviation: $\sigma_N\!=\!0.05$ and $\sigma_N\!=\!0.1$. Figs. \{(a)-(c)\} and \{(d)-(f)\} show respectively the explained variance score for the short- and long-term predictions, comparing three different structures for the TT and NN models with two activation functions (ReLU and Tanh). The results considering the training and test sets are respectively represented with darker and lighter colors.}
	\label{fig:plot_bar_MackeyGlass}
\end{figure*}

In Tables \ref{tab:mackeyglass_shortterm} and \ref{tab:mackeyglass_longterm}, we present all the results in terms of MSE, and \emph{score}, and SPCC at the convergence, for training, validation and test sets, for the short-term $x(t+6)$ and long-term $x(t+84)$ predictions respectively. All results represent the average over 400 Monte Carlo simulations, which implies 400 different random initializations. Part of these results is illustrated in Fig.  \ref{fig:plot_bar_MackeyGlass} in terms of the average \emph{score} of the training and test sets for short-term and long-term predictions. 

As expected, the performance for all models are affected with the noise addition, specially with $\sigma_{\mathrm{N}}\!=\!0.1$. According to Fig. \ref{fig:plot_bar_MackeyGlass}, the 4-th order TT ($S\!=\!3$,$R\!=\!4$) model provides the best performance for long-term predictions with the \emph{score} 0.8739, 0.8136, 0.6868 for the noiseless case, $\sigma_N\!=\!0.05$, and $\sigma_N\!=\!0.1$ respectively. However, the best performance for short-term prediction is obtained with the (4-15-1) NN with ReLU with the \emph{score} 0.8975 for $\sigma_N\!=\!0.05$ and 0.6916  for $\sigma_N\!=\!0.1$, and the 4-th order TT ($S\!=\!3$,$R\!=\!4$) with the 0.9972 for the noiseless case.

Both short-term and long-term predictions tend to provide better results, as well as the increase of coefficients. From 24/25 to 90/91 coefficients, in the best scenario, we can increase the \emph{score} until 7.23$\%$ and 6.35\% with the 4-th order TT model, 3.01\% and 1.18\% with the NN model with ReLU, and 0.38\% and 0.23\% with the NN model with Tanh, for both short-term and long-term predictions of test sets respectively. Thus, the increment of coefficients for the TT models tends to provide a bigger improvement on the test sets compared to the NN models.

\begin{figure*}[!h]
	\centering
	\includegraphics[width=.9\textwidth]{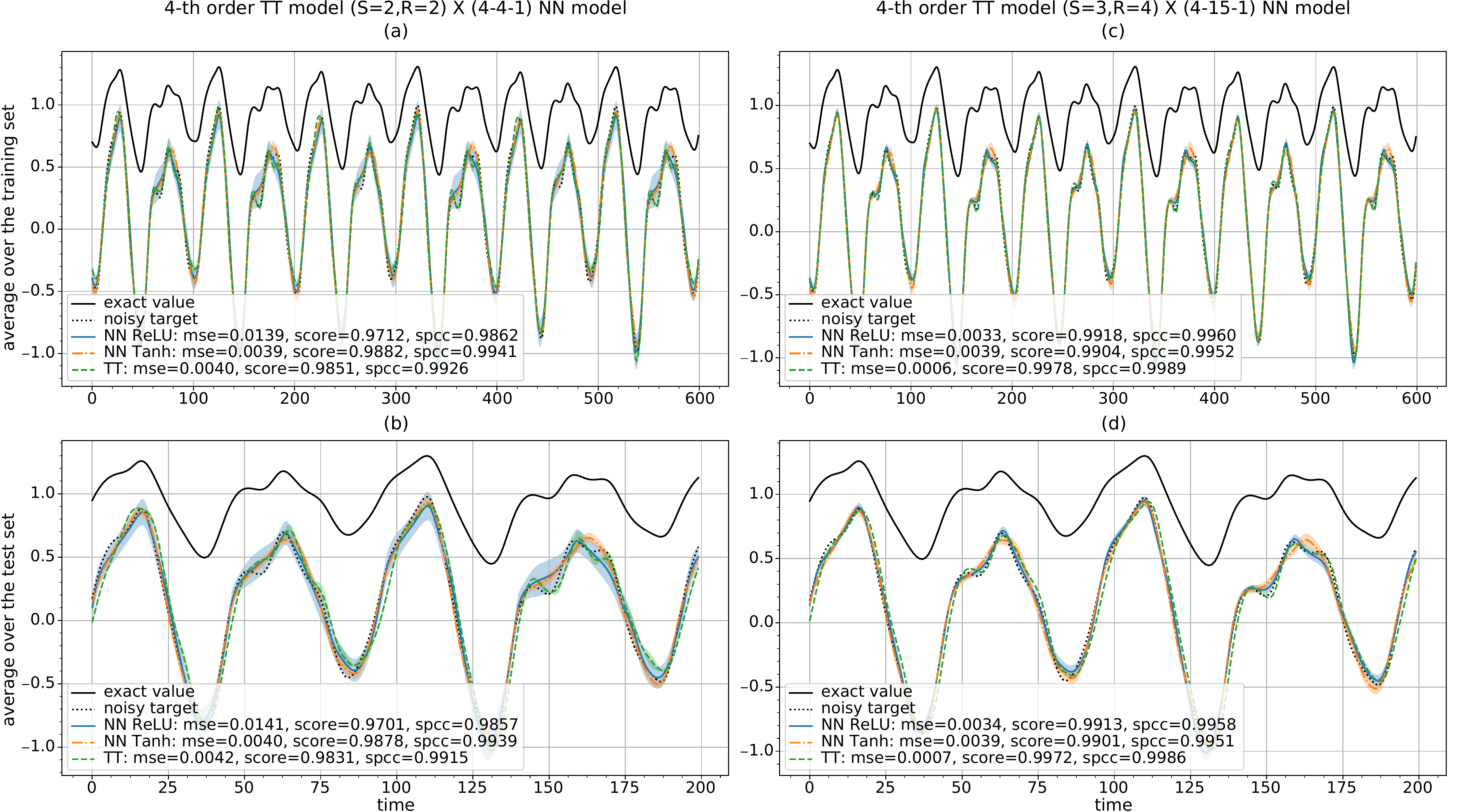}
	\caption{Short-term prediction $x(t+6)$ of Mackey-Glass time series (noiseless case) with 1000 samples (600 training + 200 validation + 200 test): average over 400 Monte Carlo simulations. Figs. (a) and (b) show the results comparing the TT model $\left(S\!=\!2,R\!=\!2\right)$ and the (4-4-1) NN model for the training and test sets respectively. Figs. (c) and (d) show the results comparing the TT model $\left(S\!=\!3,R\!=\!4\right)$ and the (4-15-1) NN model for the training and test sets respectively.}
	\label{fig:mackeyglass_timeseries_shortterm_0}
\end{figure*}

\begin{figure*}[!h]
	\centering
	\includegraphics[width=.9\textwidth]{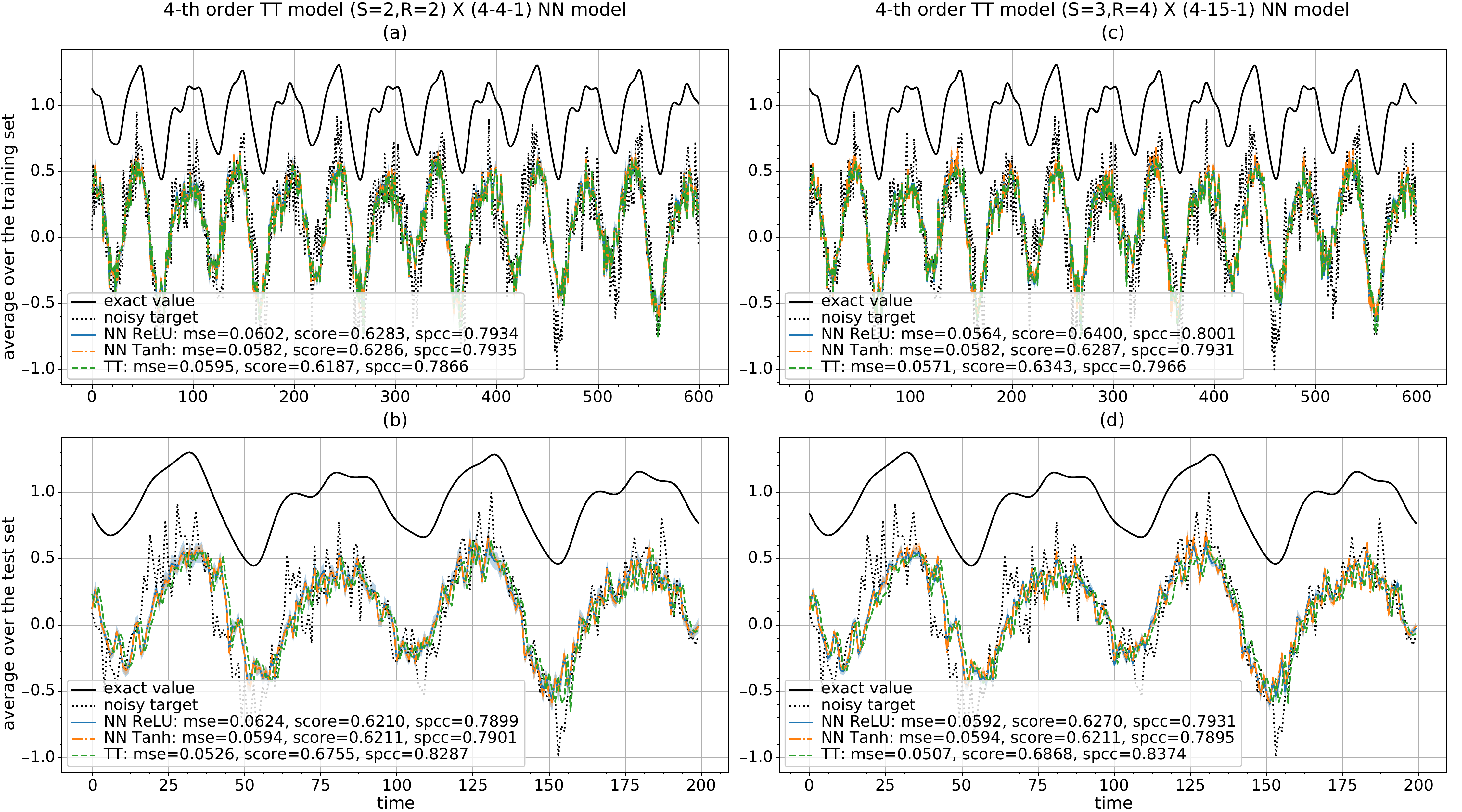}
	\caption{Long-term prediction $x(t+84)$ of Mackey-Glass time series ($\sigma_{\op{N}}=0.1$) with 1000 samples (600 training + 200 validation + 200 test): average over 400 Monte Carlo simulations. Figs. (a) and (b) show the results comparing the TT model $\left(S\!=\!2,R\!=\!2\right)$ and the (4-4-1) NN model for the training and test sets respectively. Figs. (c) and (d) show the results comparing the TT model $\left(S\!=\!3,R\!=\!4\right)$ and the (4-15-1) NN model for the training and test sets respectively.}\label{fig:mackeyglass_timeseries_longterm_01}
\end{figure*}

Figures \ref{fig:mackeyglass_timeseries_shortterm_0}-\ref{fig:mackeyglass_timeseries_longterm_01} show the amplitude versus time for the Mackey-Glass time series at the convergence, for the training and test sets, regarding the noiseless case for short-term prediction and with $\sigma_{\mathrm{N}}\!\!=\!\!0.1$ for long-term prediction respectively. The original targets (referred to in the figures as \emph{exact value}) were re-scaled into the range $[-1,1]$ and added a Gaussian noise (referred to as \emph{noisy target}) with respect to the standard deviation $\sigma_{\mathrm{N}}$. Note that each prediction curve represents the average over all Monte Carlo simulations with its respective standard deviation in order to emphasize the influence of initialization. The estimates, given by all models, tend to follow the oscillations in time of Mackey-Glass time series. The additional noise makes the forecast harder as well as the long-term predictions.

\begin{figure*}[!h]
	\centering
	\includegraphics[width=.9\textwidth]{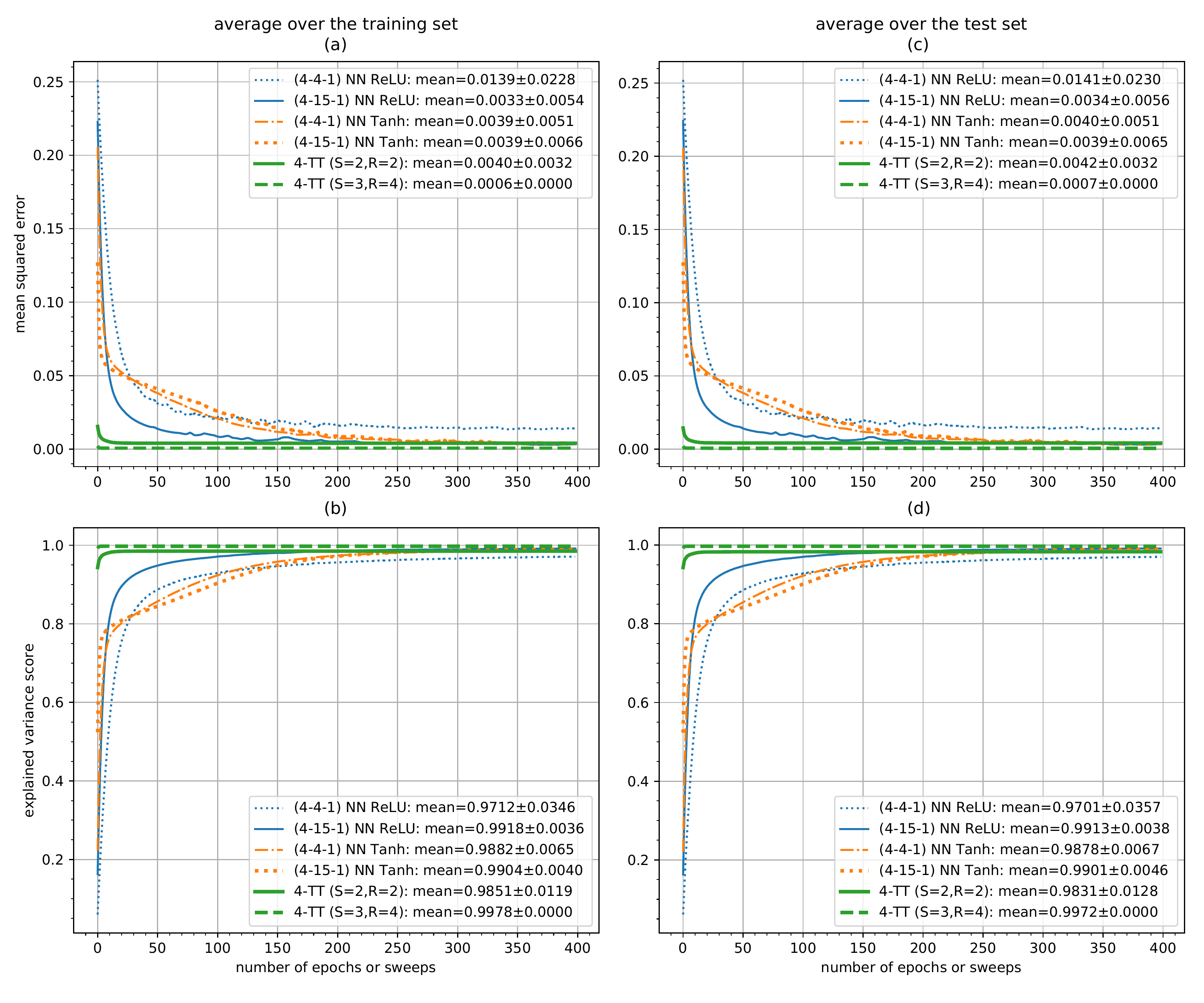}	
	\caption{Convergence analysis of Mackey-Glass time series (noiseless case), with 1000 samples (600 training + 200 validation + 200 test), for short-term prediction: average over 400 Monte Carlo simulations. Figs. \{(a),(b)\} and \{(c),(d)\} show the results comparing two TT models and four NN models for the training and test sets respectively.}\label{fig:mackeyglass_convergence_shortterm_00}
\end{figure*}

\begin{figure*}[!h]
	\centering
	\includegraphics[width=.9\textwidth]{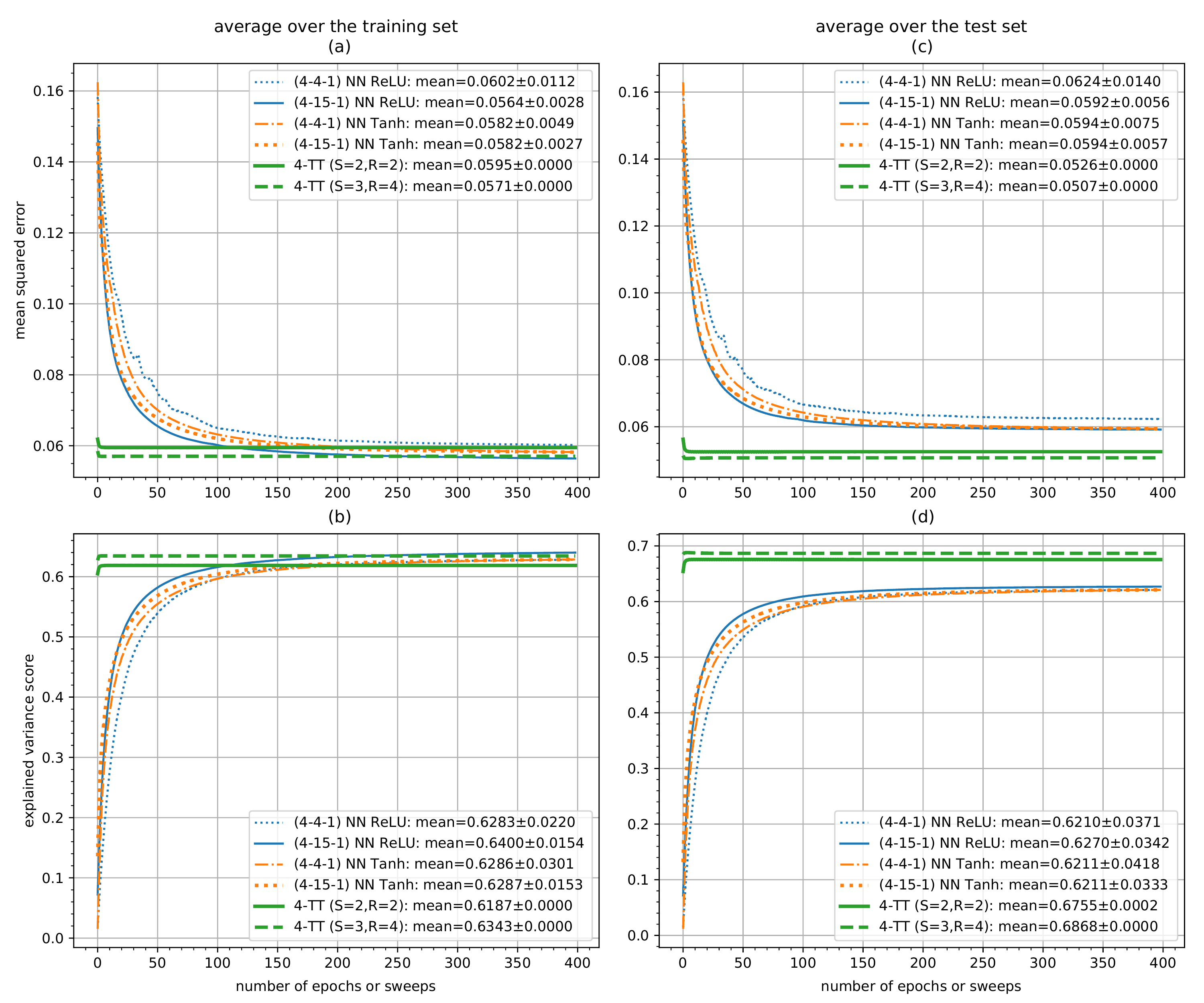}
	\caption{Convergence analysis of Mackey-Glass time series ($\sigma_{\mathrm{N}}\!\!=\!\!0.1$), with 1000 samples (600 training + 200 validation + 200 test), for long-term prediction: average over 400 Monte Carlo simulations. Figs. \{(a),(b)\}, and \{(c),(d)\} show the results comparing two TT models and four NN models for the training and test sets respectively.}\label{fig:mackeyglass_convergence_longterm_01}
\end{figure*}

The convergence of Mackey-Glass series for all configurations is represented by Figs. \ref{fig:mackeyglass_convergence_shortterm_00}-\ref{fig:mackeyglass_convergence_longterm_01}, regarding the short-term and long-term predictions, with respect to the noiseless case and $\sigma_{\mathrm{N}}\!\!=\!\!0.1$. All the curves represent the average results, in terms of MSE and \emph{score} over all Monte Carlo simulations, the mean of MSE and \emph{score} at the convergence and its respective standard deviation are denoted in the legend.

According to these figures, TT structures are faster than NN models for all configurations. We can observe that less than 10 sweeps are enough to achieve the convergence for all TT structures and, in the best case, only 2 sweeps. In contrast, NN networks with ReLU and Tanh respectively require at least 150 and 250 epochs in the best scenario.
The ReLU function provides a better convergence than Tanh, specially for short-term prediction. Furthermore, it is interesting to notice that the average performance is more representative for the TT model since the standard deviation is quite small, i.e. lower than four decimal places as indicated in the legend. Consequently, according to both figures, the initialization of coefficients in the neural networks tends to have more impact on the performance then in the tensor network, specially in the case of more coefficients and long-term predictions. 

\subsection{NASDAQ index forecasting}\label{subsec:NASDAQ}
The goal of this section is to analyze the performance of a TT network, in a real-world case, in forecasting financial time series, and compare its performance with the one obtained with the neural network model. The data were obtained from \emph{finance.yahoo.com}. The input variables of networks are given by four past values of the time series, spaced in $\Delta$ samples, which are selected through auto-correlation analysis in terms of sample Pearson correlation.

We have considered a period of a daily closing stock market of NASDAQ in USD, for short and long-term predictions, from January 2, 2018 until December 28, 2018 with $\Delta\!\!=\!\!1$ for daily predictions $x(t+1)$ and with $\Delta\!\!=\!\!30$ for monthly predictions $x(t+30)$, i.e.

\

\begin{align*}
	x(t+1)&= F\!\left(x(t-3), x(t-2), x(t-1), x(t)\right)\\
	x(t+30)&= F\!\left(x(t-90), x(t-60), x(t-30), x(t)\right).
\end{align*}
The training, validation and test sets were randomly selected from the input data and we have applied 200 Monte Carlo simulations, implying 200 different random sets with different initializations for weighting coefficients, in order to mitigate the influence of weighting initialization and the chosen sets on the algorithms. 

We apply the same procedure for selecting an optimal regularization factor, associated to the TT model,  based on the searching step, described earlier, regarding the same input interval $\{2^n\!: n\!\in\! \mathbb{Z}, -10\!\leq\! n\! \leq\! 10\}$ and considering the lowest MSE obtained from the validation set. Unlike the previous subsection, this problem requires a faster algorithm for learning NNs, with adaptive update of the learning rate; hence, we employed the Adam algorithm (originaly proposed in \cite{KiBa:14}) given in \cite{ReKaKu:18}, which is a modified version without the debiased step of the original version, with the following hyper-parameters, typically recommended in practice \cite{KiBa:14,ReKaKu:18}: the initial learning rate $\alpha\!\!=\!\!0.001$ with the exponential decay rates for the first and second moment estimates $\beta_1\!=\!0.9$ and $\beta_2\!=\!0.99$, a small number to prevent any division by zero in the implementation $\epsilon\!=\!10^{-8}$.

Five different structures have been chosen for the TT and NN models and employed to predict the short and long-term indices, with approximate number of coefficients, i.e. (24, 90, 180, 544, 1300) and (25, 91, 181, 547, 1303) for both respective structures. For the TT model, we have: ($S\!\!=\!\!2$,$R\!\!=\!\!2$), ($S\!\!=\!\!3$,$R\!\!=\!\!4$), ($S\!\!=\!\!3$,$R\!\!=\!\!9$), ($S\!\!=\!\!4$,$R\!\!=\!\!16$), ($S\!\!=\!\!5$,$R\!\!=\!\!25$). For the NN model, we have: (4-4-1), (4-15-1), (4-30-1), (4-91-1), (4-217-1) with two activation functions Tanh and ReLU.

\begin{figure*}[!h]
	\centering
	\includegraphics[width=.9\textwidth]{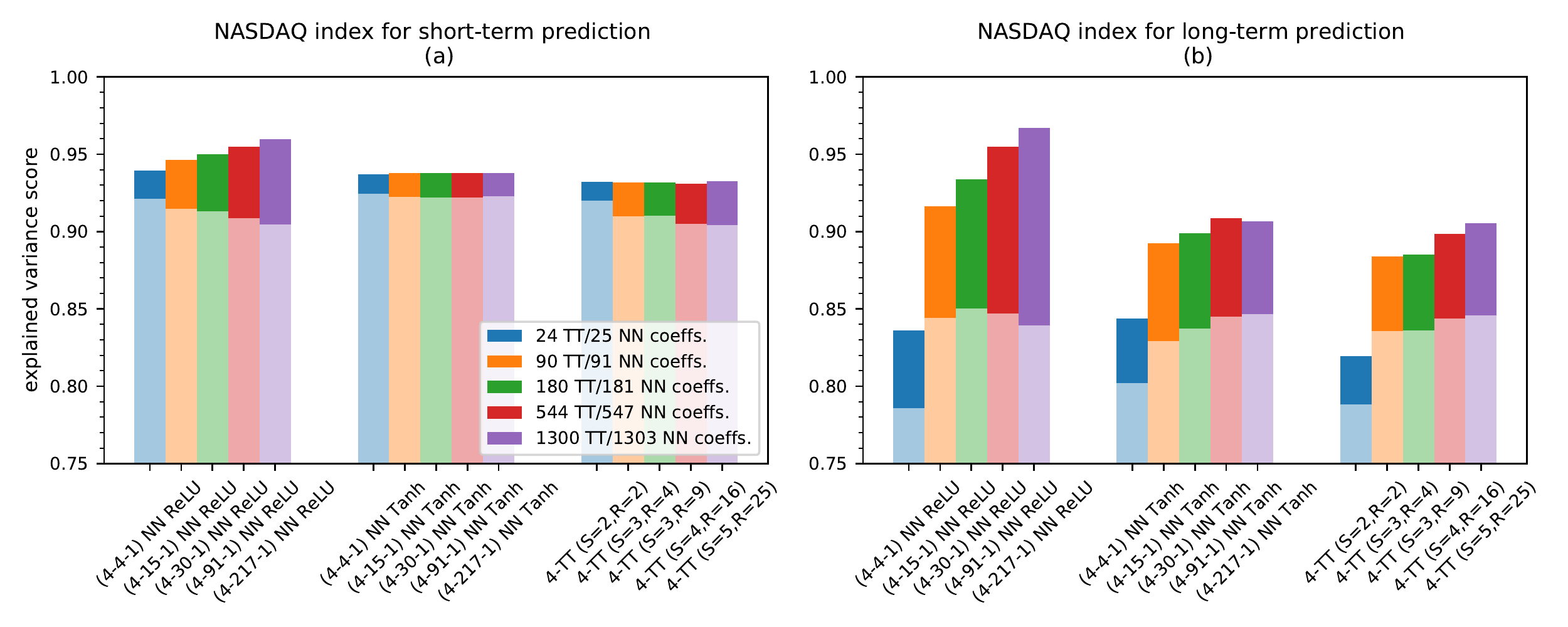}
	\caption{NASDAQ index forecasting: Figs. (a) and (b) show respectively the explained variance score for the short- and long-term predictions, comparing five different structures for the TT and NN models with two activation functions (ReLU and Tanh). The results considering the training and test sets are respectively represented with darker and lighter colors.}
	\label{fig:plot_bar_NASDAQ}
\end{figure*}

In Tables \ref{tab:nasdaq_shortterm} and \ref{tab:nasdaq_longterm}, all results are shown in terms of MSE, \emph{score}, SPCC, and \emph{$R$-squared} at the convergence, for training, validation, and test sets, for the short-term $x(t+1)$ and long-term $x(t+30)$ predictions respectively. Part of these results is illustrated in Fig. \ref{fig:plot_bar_NASDAQ}
in terms of the average \emph{score} of the training and test sets for short-term and long-term predictions. According to Fig. \ref{fig:plot_bar_NASDAQ}, we can note that the performance of both models for the daily prediction does not have a significant improvement on the training set with the increase of coefficients, from 25/24 to 1303/1300, mainly for the TT and NN model with Tanh function, lower than two decimal places. 

Furthermore, it is possible to check a decrement on the performance of training and test sets when more coefficients are employed, regarding the average \emph{score} respectively of the validation and test sets, of 1.67\% and 1.81\% for the NN with ReLU, 0.11\% and 0.16\% for the NN with Tanh, and 0.78\% and 1.72\% for the TT model. These decays indicate a tendency to over-fitting of all models: thus, more coefficients will not provide better results associated to the test set. The best performance regarding the test sets is obtained with the (4-4-1) NN model with Tanh with the \emph{score} 0.9243, followed by the (4-4-1) NN with ReLU with 0.9212 and the 4-th order TT model with ($S\!\!=\!\!2$,$R\!\!=\!\!2$) with 0.9200, respectively representing a reduction of 0.34\% and 0.46\% with respect to the best \emph{score}.

In contrast, taking into account Table \ref{tab:nasdaq_longterm}, we verify a simultaneous improvement for the monthly predictions on the training, validation, and test sets, except to the NN model with ReLU. For this last structure, we observe a decay of the performance on the validation and test sets when we employ more than 30 hidden neurons in the intermediate layer. Therefore, the best result is achieved with the highest number of coefficients only with the NN model with Tanh and the TT model. The (4-217-1) NN model with Tanh, the 4-th order TT model with ($S\!\!=\!\!5$,$R\!\!=\!\!25$), and the (4-30-1) NN model with ReLU respectively provide a \emph{score} 0.8465, 0.8458 and 0.8501, which represent an increment on the test set of 5.54\%, 7.29\%, 8.21\% respectively with respect to the worst configuration, i.e. the case with the lowest number of coefficients for each model. Note that this improvement was achieved by the increase of coefficients, from 24/25/25 to 1303/181/1300 coefficients for respectively the TT, the NN with ReLU and the NN with Tanh models. Therefore, both TT and NN with Tanh provide similar performances, but the TT showed a higher increment on the performance of test sets when more coefficients are considered.

\begin{figure*}[!h]
	\centering
	\includegraphics[width=.9\textwidth]{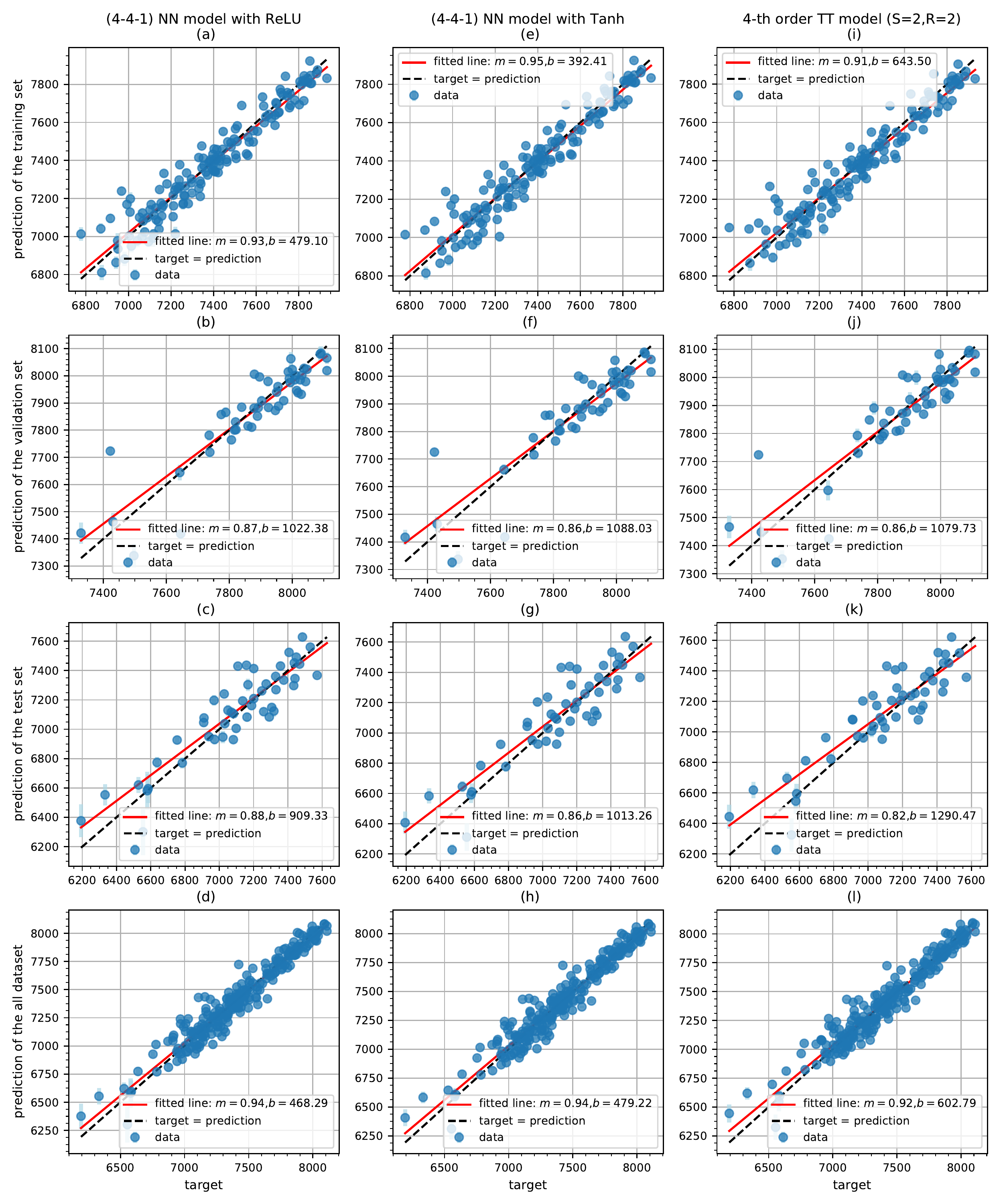}	
	\caption{Short-term prediction $x(t+1)$ of NASDAQ index with 246 samples (148 training + 49 validation + 49 test): average over 200 Monte Carlo simulations. Figs. \{(a)-(d)\}, \{(e)-(h)\} and \{(i)-(l)\} show respectively the results for the (4-4-1) NN model with ReLU, the (4-4-1) NN model with Tanh, and the TT model $\left(S\!=\!2,R\!=\!2\right)$.}\label{fig:finance_fitline_shortterm}
\end{figure*}

\begin{figure*}[!h]
	\centering
	\includegraphics[width=.9\textwidth]{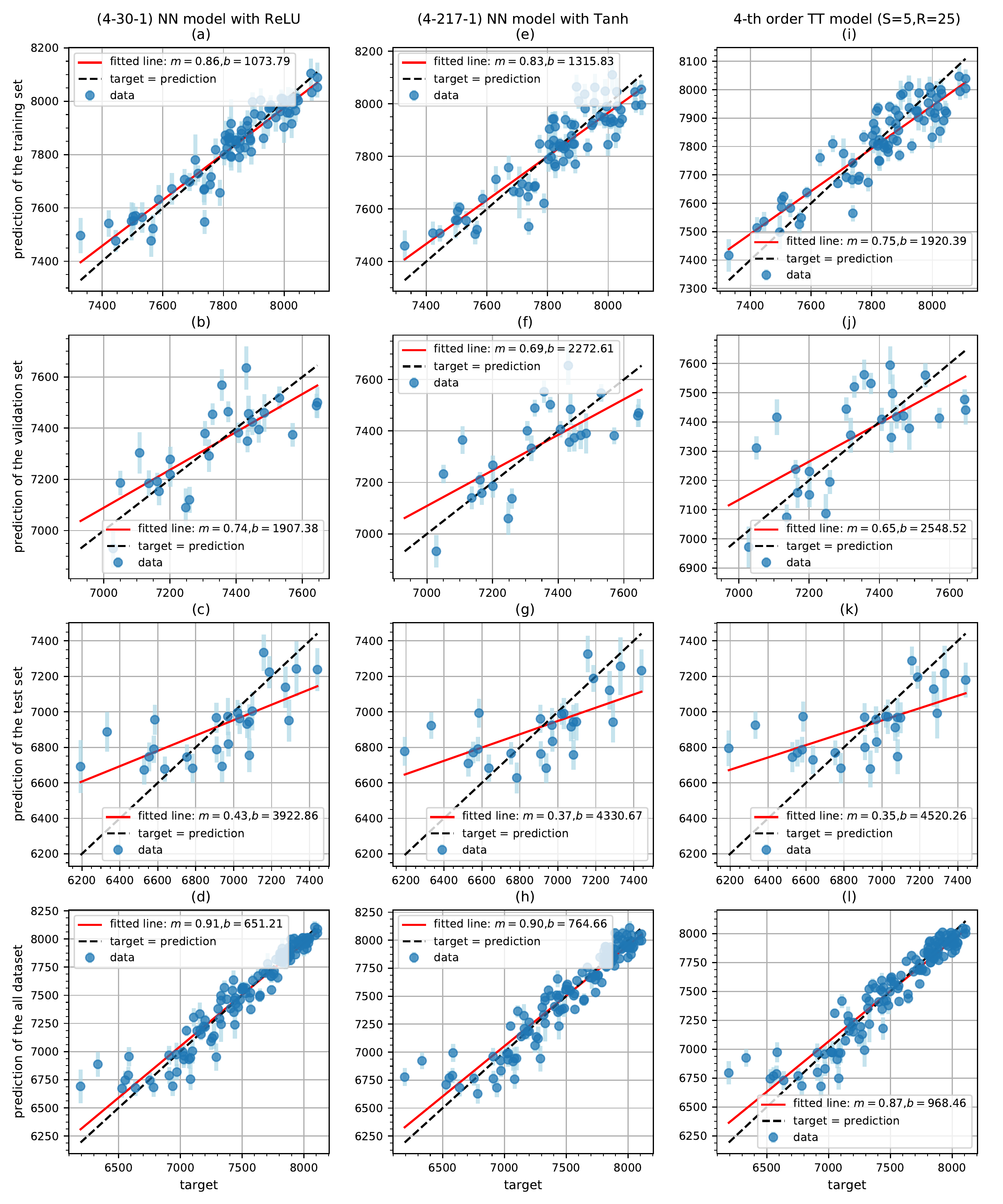}
	\caption{Long-term prediction $x(t+30)$ of NASDAQ index with 130 samples (78 training + 26 validation + 26 test): average over 200 Monte Carlo simulations. Figs. \{(a)-(d)\}, \{(e)-(h)\} and \{(i)-(l)\} show respectively the results for the (4-30-1) NN model with ReLU, the (4-217-1) NN model with Tanh, and the TT model $\left(S\!=\!5,R\!=\!25\right)$.}\label{fig:finance_fitline_longterm}
\end{figure*}

Figures \ref{fig:finance_fitline_shortterm} and \ref{fig:finance_fitline_longterm} represent the relation between the short-term $\hat x(t+1)$ and long-term $\hat x(t+30)$ predictions, with the respective standard deviations, versus the desired target ($x(t+1)$ or $x(t+30)$) by separately taking into account the predictions of the training, validation, test and all sets for each model. The best-fitting (or regression) line and the fitted line associated to each prediction, through the slope $m$ and the $y$-intercept $b$ of each red line, are indicated in each chart. Note that only the best configuration for each model is presented in this figure, for daily and monthly predictions, as discussed above. It is important to emphasize that this kind of chart presents a visualization resource for the learned predictions and it will not necessarily point out the same best model since the best-fitting line is given by a straight line, which linearly maps the error of predictions.

When the prediction is closer enough to the desired value, the slope tends to one as well as the $y$-intercept tends to zero, thus, in the ideal case, we have $m\!\!\approx\!\!1$ and $b\!\approx\!0$. From these figures, we verify, as expected, that the predictions of the training set (even as all data sets) provide better fitting performance once both ideal and fitted lines are closer than the lines associated to the predictions of the validation and test sets. Furthermore, as also expected, we obtain worse performances for monthly predictions than the daily predictions. 

Figure \ref{fig:finance_fitline_shortterm} indicates the best fit of slope and $y$-intercept achieved for the test set with the NN model with ReLU ($m\!=\!0.88$,$b\!=\!909.33$), followed by the NN with Tanh with ($m\!=\!0.86$,$b\!=\!1013.26$) and the TT model with ($m\!=\!0.82$, $b\!=\!1290.47$). According to Fig. \ref{fig:finance_fitline_longterm}, by considering long-term predictions, the best learned slope and $y$-intercept for the test set is achieved with the NN model with ReLU with ($m\!=\!0.43$, $b\!=\!3922.86$), followed by the NN with Tanh with ($m\!=\!0.37$, $b\!=\!4330.67$) and the TT model with ($m\!=\!0.35$, $b\!=\!4520.26$).

\begin{figure*}[!h]
	\centering
	\includegraphics[width=.9\textwidth]{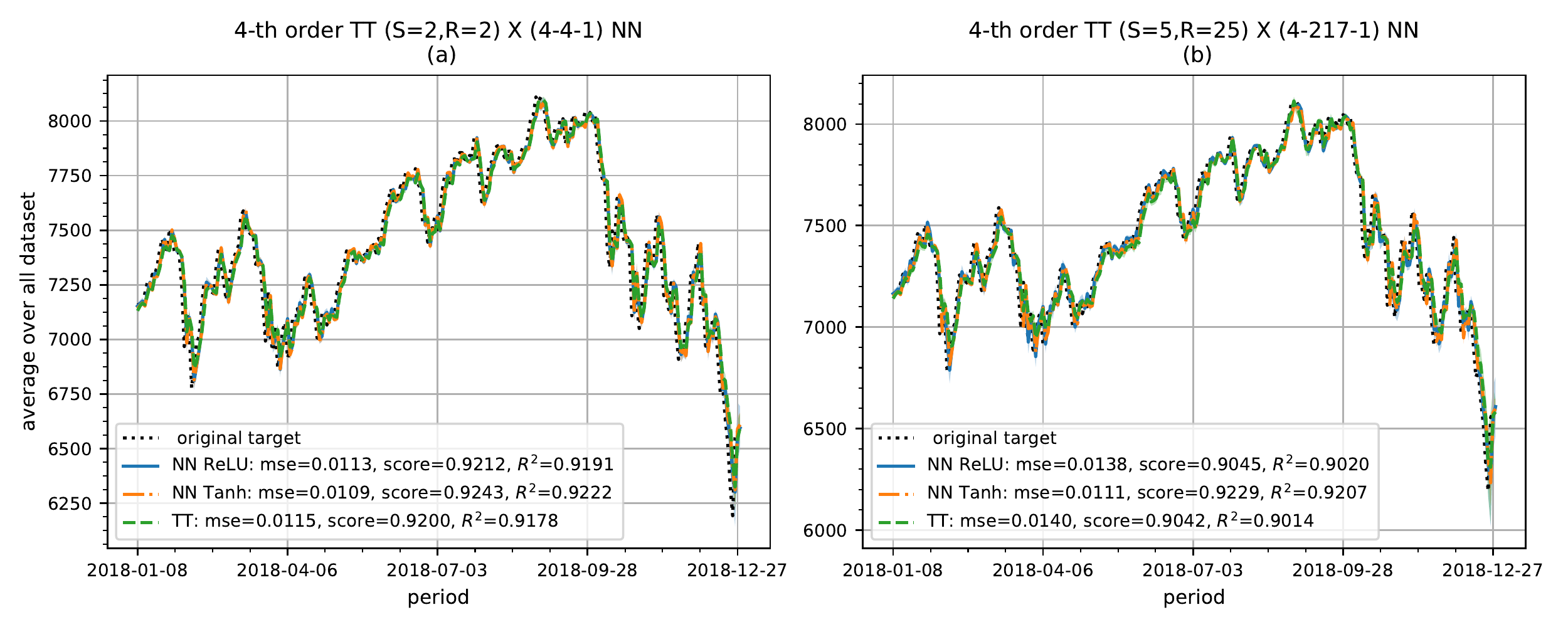}
	\caption{Short-term prediction $x(t+1)$ of NASDAQ index with 246 samples (148 training + 49 validation + 49 test): average over 200 Monte Carlo simulations. Figs. (a) and (b) show respectively the results comparing the TT model $\left(S\!=\!2,R\!=\!2\right)$ and the (4-4-1) NN model with Tanh and ReLU, and the TT model $\left(S\!=\!5,R\!=\!25\right)$ and the (4-217-1) NN model with Tanh and ReLU.}
	\label{fig:finance_timeseries_shortterm}
\end{figure*}

\begin{figure*}[!h]
	\centering
	\includegraphics[width=.9\textwidth]{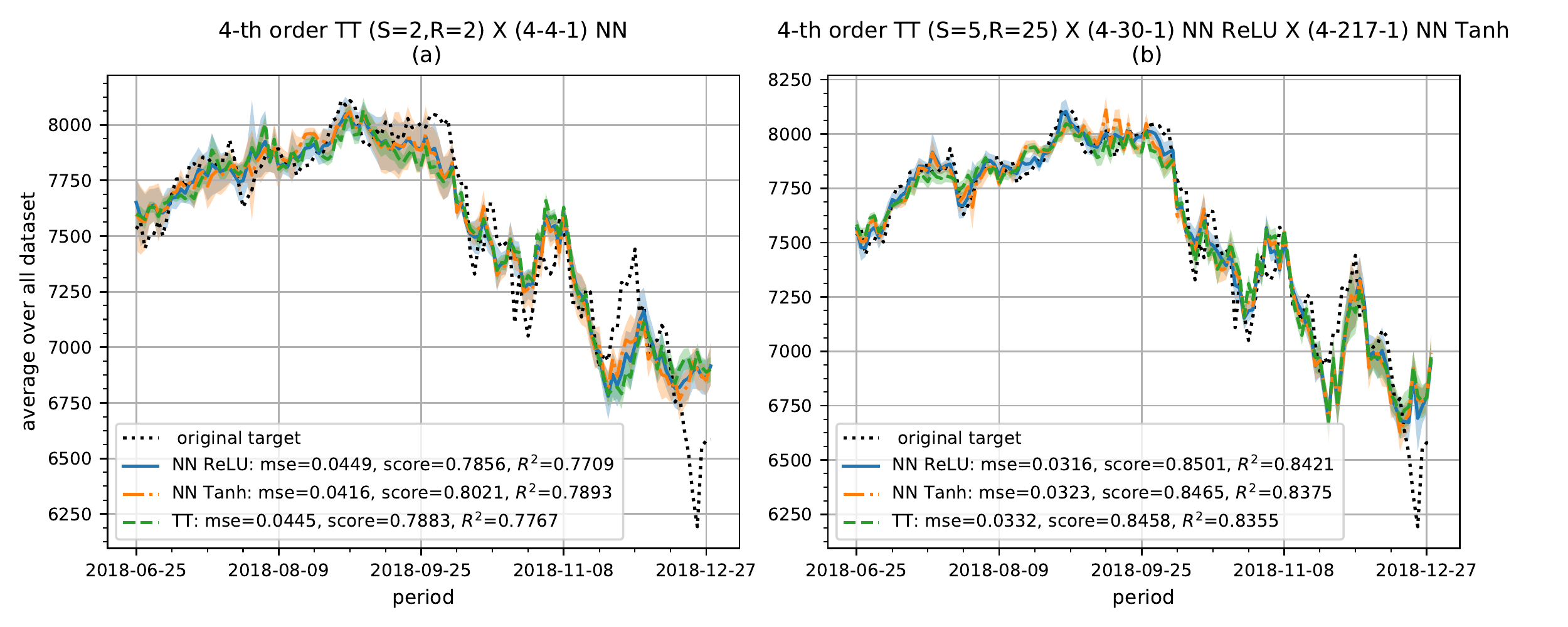}
	\caption{Long-term prediction $x(t+30)$ of NASDAQ index with 130 samples (78 training + 26 validation + 26 test): average over 200 Monte Carlo simulations. Figs. (a) and (b) show the results comparing the TT model $\left(S\!=\!2,R\!=\!2\right)$ and the (4-4-1) NN model with Tanh and ReLU, and the TT model $\left(S\!=\!5,R\!=\!25\right)$, the (4-30-1) NN model with ReLU, and the (4-217-1) NN model with Tanh respectively.}
	\label{fig:finance_timeseries_longterm}
\end{figure*}

The NASDAQ index predictions of all data set over the selected time period at the convergence for both short- and long-term predictions are presented in Figs. \ref{fig:finance_timeseries_shortterm}-\ref{fig:finance_timeseries_longterm}, for only two different configurations for each structure. The original target is also represented in these figures as well as the average MSE, \emph{score} and \emph{$R$-squared} of test sets over all Monte Carlo simulations. Comparing both figures, observe that the standard deviation of the predictions are more visible for monthly predictions, i.e. $x\left(t\!+\!30\right)$, than for daily predictions $x\left(t\!+\!1\right)$. Despite the difference of performance between all models, we can observe that the learned models follow the oscillations of the index in time, mainly for daily forecast. Furthermore, in Figure \ref{fig:finance_timeseries_longterm}, we can note a visual difference between both predictions with 24/25/25 and 1303/181/1300 coefficients for respectively the TT/ NN with ReLU and NN with Tanh models, unlike the short-term predictions. 

\begin{figure*}[!h]
	\centering
	\includegraphics[width=.9\textwidth]{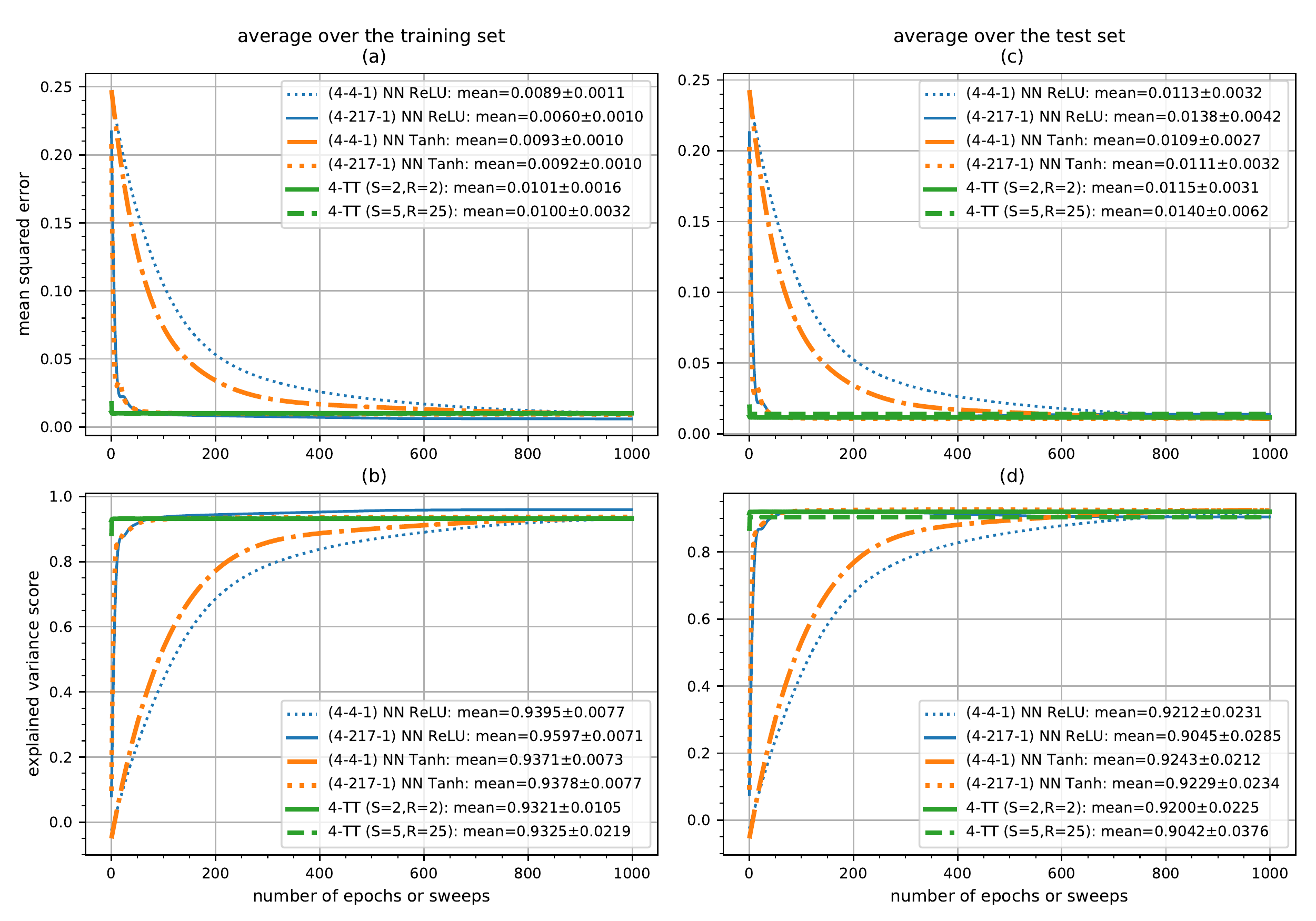}
	\caption{Convergence analysis of NASDAQ index, with 246 samples (148 training + 49 validation + 49 test), for short-term prediction: average over 200 Monte Carlo simulations. Figs. \{(a),(b)\} and \{(c),(d)\} show the results comparing two TT models and four NN models for the training and test sets respectively.}
\label{fig:finance_convergence_shortterm}
\end{figure*}

\begin{figure*}[!h]
	\centering
	\includegraphics[width=.9\textwidth]{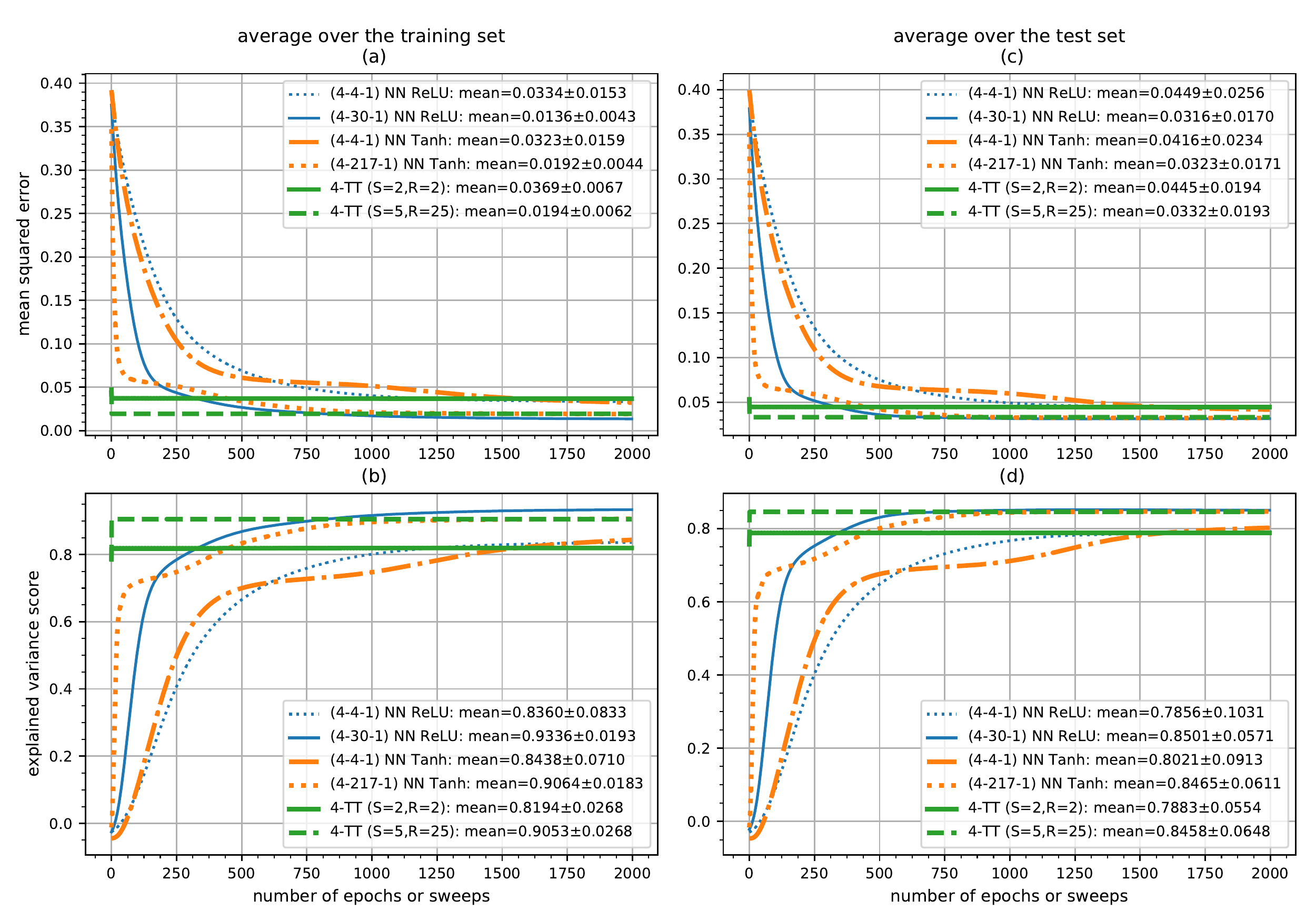}	
	\caption{Convergence analysis of NASDAQ index, with 130 samples (78 training + 26 validation + 26 test), for long-term prediction: average over 200 Monte Carlo simulations. Figs. \{(a),(b)\} and \{(c),(d)\} show the results comparing two TT models and four NN models for the training and test sets respectively.}\label{fig:finance_convergence_longterm}
\end{figure*}

Figures \ref{fig:finance_convergence_shortterm} and \ref{fig:finance_convergence_longterm} show the convergence of NASDAQ index forecasting of short- and long-term for the training and test sets. The averages of MSE and \emph{score} over all Monte Carlos simulation are shown in these figures and we denote the mean of MSE and \emph{score} at the convergence and its respective standard deviation in the legend. Clearly, the TT models present the fastest convergence, the maximum of 6 sweeps is required; on the other hand, the NN models with ReLU and Tanh require more 2000 epochs in the worst scenario. It is interesting to observe that the standard deviations, denoted in the figures, shown the proximity of the results at the convergence and the influence of random initialization of weighting networks and the selection of the datasets.

\section{Conclusions}\label{sec:conclusions}
A key feature of this article is to analyze the ability of the use of TT networks as an efficient tool to compress MLPs weights, independently of a particular application. Expressions were derived in order to view the non-linear global problem for estimating tensor structure as a set of several sub-problems for estimating each core by means of a closed-form solution based on the conventional solution for a general regression model. Furthermore, these expressions provide a reduction of required memory and computational complexity. The addition of a matrix regularization factor in the loss function enables a parameter to adjust the model flexibility for the purpose of providing a balance between training performance and capability of the model generalization, i.e. by preventing over-fitting problem.

From the first part of our simulations, MLPs were modeled by TT networks, which enabled a powerful compressed representation of a simple MLP in terms of the number of coefficients with similar performance. The smallest adopted TT network with the lowest number of coefficients, representing a reduction of 95\% of NN coefficients, provided an average \emph{score} over the test set equal to 0.8110 and 0.8958 depending on the adopted activation function (i.e. Tanh and ReLU functions respectively). The best \emph{score}, achieved with 20\% of reduction in terms of number coefficients, is 0.9101 and 0.9880 for both Tanh and ReLU functions respectively. Furthermore, we verified the relevance of non-linearity introduced by feature mapping, which can enable a better model fitting with lower number of coefficients.

The second part was driven by applications in noisy chaotic time series and stock market index price forecasting, by means of Mackey-Glass equation and NASDAQ index. The estimates, given by neural and tensor networks, tend to follow the oscillations in time of Mackey-Glass time series. As expected, the additional noise makes the forecast harder as well as the long-term predictions. TT structures tended to provide better performances over test sets regarding networks with more coefficients. In addition, the increment of coefficients for the TT models tends to provide a bigger improvement on the test sets compared to the NN models. Besides that we have observed that the initialization of coefficients in the neural networks tends to have more impact on the performance then in the tensor network, specially in the case of more coefficients and long-term predictions.

From the results regarding the NASDAQ index forecasting, differently to the previous time series, we have noted a tendency to over-fitting of all models, mainly for daily predictions. The performance of both NN and TT models over test sets, for the daily prediction, does not improve with the increase of coefficients. In contrast, we verify a simultaneous improvement for the monthly predictions on the training, validation, and test sets, except to the NN model with ReLU. Both models, the NN with Tanh and TT models, provide the similar performance, however the TT showed a higher increment on the performance over test sets when more coefficients are considered.

In terms of convergence speed, tensor networks tend to achieve faster convergence, thanks to the closed-form solution. We also observed that neural networks are quite sensitive to the adjustment of hyper-parameters and may require more sophisticated adaptive learning-method algorithms for solving more complicated problems in order to accelerate the convergence. When we consider more  sophisticated methods, the algorithms tend to be more robust, on the other hand, more hyper-parameters will be probably required. 

By concluding, tensor networks are promising to design architectures of the DNNs more efficiently, and also they can accelerate and simplify the learning process in terms of network structure, algorithm tuning procedure, computational complexity and required memory, while maintaining a reasonable quality of prediction.

\appendix
\section{}
\vspace{-5cm}

\begin{landscape}
\begin{table*}
\centering
\caption{Recovering the (10-200-1) neural network with Tanh function}\label{tab:recoverNN_tanh}
{\scriptsize{\begin{tabular}{c c c c c c c c c c c c c c c}
    \toprule
     & \multirow{2}{*}{$R$} & {\textbf{no. of}} & \multicolumn{4}{c}{\textbf{training}} & \multicolumn{4}{c}{\textbf{validation}} & \multicolumn{4}{c}{\textbf{test}}\\
    \cmidrule{4-15} 
    & & {\textbf{coeffs.}} & \text{MSE} & \text{\emph{score}} & \text{SPCC} & \text{\emph{$R$-squared}} & \text{MSE} & \text{\emph{score}} & \text{SPCC} & \text{\emph{$R$-squared}} & \text{MSE} & \text{\emph{score}} & \text{SPCC} & \text{\emph{$R$-squared}}\\
    \midrule
    \parbox[t]{2mm}{\multirow{9}{*}{\rotatebox[origin=c]{90}{$S$=2}}} & 4 & 232 & 9.922e+01 & 0.8418 & 0.9176 & 0.8418 & 1.036e+02 & 0.8354 & 0.9140 & 0.8352 & 1.036e+02 & 0.8345 & 0.9136 & 0.8344\\
     & 6 & 424 & 8.269e+01 & 0.8682 & 0.9319 & 0.8682 & 8.918e+01 & 0.8583 & 0.9265 & 0.8582 & 8.899e+01 & 0.8578 & 0.9262 & 0.8577\\
     & 8 & 680 & 7.393e+01 & 0.8822 & 0.9394 & 0.8822 & 8.199e+01 & 0.8698 & 0.9326 & 0.8696 & 8.178e+01 & 0.8693 & 0.9324 & 0.8692\\     
     & 10 & 888 & 7.033e+01 & 0.8879 & 0.9424 & 0.8879 & 7.969e+01 & 0.8734 & 0.9346 & 0.8733 & 7.958e+01 & 0.8728 & 0.9343 & 0.8728\\     
     & 12 & 1128 & 6.861e+01 & 0.8906 & 0.9440 & 0.8906 & 7.890e+01 & 0.8746 & 0.9352 & 0.8746 & 7.883e+01 & 0.8740 & 0.9350 & 0.8740\\
     & 14 & 1400 & 6.723e+01 & 0.8928 & 0.9451 & 0.8928 & 7.842e+01 & 0.8754 & 0.9357 & 0.8753 & 7.831e+01 & 0.8749 & 0.9354 & 0.8748\\                    
     & 20 & 1960 & 6.620e+01 & 0.8945 & 0.9460 & 0.8945 & 7.812e+01 & 0.8759 & 0.9359 & 0.8758 & 7.803e+01 & 0.8753 & 0.9356 & 0.8752\\
     & 25 & 2280 & 6.604e+01 & 0.8947 & 0.9462 & 0.8947 & 7.809e+01 & 0.8759 & 0.9359 & 0.8758 & 7.802e+01 & 0.8753 & 0.9356 & 0.8752\\          
     & 30 & 2600 & 6.602e+01 & 0.8948 & 0.9462 & 0.8948 & 7.807e+01 & 0.8760 & 0.9360 & 0.8759 & 7.795e+01 & 0.8754 & 0.9357 & 0.8754\\     
   & \color{red}40 & \color{red}2728 & \color{red}6.586e+01 & \color{red}0.8950 & \color{red}0.9463 & \color{red}0.8950 & \color{red}7.796e+01 & \color{red}0.8762 & \color{red}0.9361 & \color{red}0.8761 & \color{red}7.785e+01 & \color{red}0.8756 & \color{red}0.9358 & \color{red}0.8755\\
    \midrule
    \parbox[t]{2mm}{\multirow{7}{*}{\rotatebox[origin=c]{90}{$S$=3}}} & 2 & 108 & 1.156e+02 & 0.8156 & 0.9031 & 0.8156 & 1.186e+02 & 0.8115 & 0.9009 & 0.8114 & 1.182e+02 & 0.8110 & 0.9007 & 0.8109\\    
     & 4 & 378 & 7.740e+01 & 0.8766 & 0.9363 & 0.8766 & 8.413e+01 & 0.8663 & 0.9308 & 0.8662 & 8.458e+01 & 0.8648 & 0.9300 & 0.8647\\     
     & 6 & 774 & 5.785e+01 & 0.9078 & 0.9528 & 0.9078 & 6.827e+01 & 0.8915 & 0.9442 & 0.8915 & 6.840e+01 & 0.8907 & 0.9438 & 0.8906\\     
     & 8 & 1314 & 4.673e+01 & 0.9255 & 0.9621 & 0.9255 & 5.974e+01 & 0.9051 & 0.9514 & 0.9050 & 5.996e+01 & 0.9042 & 0.9509 & 0.9041\\     
     & 10 & 1920 & 4.066e+01 & 0.9352 & 0.9672 & 0.9352 & 5.614e+01 & 0.9108 & 0.9544 & 0.9107 & 5.625e+01 & 0.9101 & 0.9540 & 0.9101\\
     & \color{red}12 & \color{red}2556 & \color{red}3.733e+01 & \color{red}0.9405 & \color{red}0.9699 & \color{red}0.9405 & \color{red}5.465e+01 & \color{red}0.9132 & \color{red}0.9556 & \color{red}0.9131 & \color{red}5.467e+01 & \color{red}0.9126 & \color{red}0.9554 & \color{red}0.9126\\     
     & 14 & 3288 & 3.597e+01 & 0.9427 & 0.9711 & 0.9427 & 5.511e+01 & 0.9124 & 0.9552 & 0.9124 & 5.509e+01 & 0.9120 & 0.9550 & 0.9119\\
    \bottomrule
\end{tabular}}}
\end{table*}

\begin{table*}
\centering
\caption{Recovering the (10-200-1) neural network with ReLU function}\label{tab:recoverNN_relu}
{\scriptsize{\begin{tabular}{c c c c c c c c c c c c c c c}
    \toprule
     & \multirow{2}{*}{$R$} & {\textbf{no. of}} & \multicolumn{4}{c}{\textbf{training}} & \multicolumn{4}{c}{\textbf{validation}} & \multicolumn{4}{c}{\textbf{test}}\\
    \cmidrule{4-15} 
    & & {\textbf{coeffs.}} & \text{MSE} & \text{\emph{score}} & \text{SPCC} & \text{\emph{$R$-squared}} & \text{MSE} & \text{\emph{score}} & \text{SPCC} & \text{\emph{$R$-squared}} & \text{MSE} & \text{\emph{score}} & \text{SPCC} & \text{\emph{$R$-squared}}\\
    \midrule
    \parbox[t]{2mm}{\multirow{9}{*}{\rotatebox[origin=c]{90}{$S$=2}}} & 4 & 232 & 1.858e+02 & 0.9550 & 0.9772 & 0.9550 & 1.932e+02 & 0.9533 & 0.9764 & 0.9533 & 1.928e+02 & 0.9533 & 0.9764 & 0.9533\\
     & 6 & 424 & 1.414e+02 & 0.9657 & 0.9827 & 0.9657 & 1.527e+02 & 0.9631 & 0.9814 & 0.9631 & 1.534e+02 & 0.9629 & 0.9813 & 0.9628\\     
     & 8 & 680 & 1.307e+02 & 0.9683 & 0.9841 & 0.9683 & 1.452e+02 & 0.9649 & 0.9823 & 0.9649 & 1.462e+02 & 0.9646 & 0.9822 & 0.9646\\     
     & 10 & 888 & 1.262e+02 & 0.9694 & 0.9846 & 0.9694 & 1.439e+02 & 0.9652 & 0.9825 & 0.9652 & 1.448e+02 & 0.9650 & 0.9824 & 0.9649\\          
     & 12 & 1128 & 1.236e+02 & 0.9700 & 0.9849 & 0.9700 & 1.437e+02 & 0.9653 & 0.9825 & 0.9653 & 1.444e+02 & 0.9650 & 0.9824 & 0.9650\\
     & 14 & 1400 & 1.219e+02 & 0.9704 & 0.9852 & 0.9704 & 1.435e+02 & 0.9653 & 0.9825 & 0.9653 & 1.440e+02 & 0.9652 & 0.9824 & 0.9651\\          
     & \color{red}20 & \color{red}1960 & \color{red}1.201e+02 & \color{red}0.9709 & \color{red}0.9854 & \color{red}0.9709 & \color{red}1.434e+02 & \color{red}0.9654 & \color{red}0.9825 & \color{red}0.9653 & \color{red}1.438e+02 & \color{red}0.9652 & \color{red}0.9825 & \color{red}0.9652\\     
     & 25 & 2280 & 1.196e+02 & 0.9710 & 0.9854 & 0.9710 & 1.433e+02 & 0.9654 & 0.9826 & 0.9654 & 1.436e+02 & 0.9652 & 0.9825 & 0.9652\\         
     & 30 & 2600 & 1.195e+02 & 0.9710 & 0.9854 & 0.9710 & 1.433e+02 & 0.9654 & 0.9826 & 0.9654 & 1.436e+02 & 0.9652 & 0.9825 & 0.9652\\ 
    \midrule
    \parbox[t]{2mm}{\multirow{7}{*}{\rotatebox[origin=c]{90}{$S$=3}}} & 2 & 108 & 4.193e+02 & 0.8984 & 0.9478 & 0.8984 & 4.289e+02 & 0.8964 & 0.9468 & 0.8963 & 4.304e+02 & 0.8958 & 0.9465 & 0.8958\\       
     & 4 & 378 & 1.070e+02 & 0.9741 & 0.9870 & 0.9741 & 1.179e+02 & 0.9715 & 0.9857 & 0.9715 & 1.168e+02 & 0.9717 & 0.9858 & 0.9717\\          
     & 6 & 774 & 5.379e+01 & 0.9870 & 0.9935 & 0.9870 & 6.534e+01 & 0.9842 & 0.9921 & 0.9842 & 6.536e+01 & 0.9842 & 0.9921 & 0.9842\\          
     & 8 & 1314 & 4.022e+01 & 0.9902 & 0.9951 & 0.9902 & 5.322e+01 & 0.9871 & 0.9936 & 0.9871 & 5.324e+01 & 0.9871 & 0.9935 & 0.9871\\
     & 10 & 1920 & 3.338e+01 & 0.9919 & 0.9960 & 0.9919 & 4.966e+01 & 0.9880 & 0.9940 & 0.9880 & 4.950e+01 & 0.9880 & 0.9940 & 0.9880\\          
     & \color{red}12 & \color{red}2556 & \color{red}2.927e+01 & \color{red}0.9929 & \color{red}0.9965 & \color{red}0.9929 & \color{red}4.772e+01 & \color{red}0.9885 & \color{red}0.9942 & \color{red}0.9885 & \color{red}4.775e+01 & \color{red}0.9884 & \color{red}0.9942 & \color{red}0.9884\\     
     & 14 & 3288 & 2.945e+01 & 0.9929 & 0.9964 & 0.9929 & 5.019e+01 & 0.9879 & 0.9939 & 0.9878 & 4.989e+01 & 0.9879 & 0.9939 & 0.9879\\
    \bottomrule
\end{tabular}}}
\end{table*}
\end{landscape}


\begin{table*}[!h]
\centering
\caption{Mackey-Glass time series for short-term prediction}\label{tab:mackeyglass_shortterm}
{\scriptsize{\begin{tabular}{c c c c c c c c c c c}
    \toprule
    \multirow{2}{*}{\textbf{models}} & \multirow{2}{*}{$\sigma_{\mathrm{N}}$} & \multicolumn{3}{c}{\textbf{training}} & \multicolumn{3}{c}{\textbf{validation}} & \multicolumn{3}{c}{\textbf{test}}\\
    \cmidrule{3-11} 
    & & \text{MSE} & \text{\emph{score}} & \text{SPCC} & \text{MSE} & \text{\emph{score}} & \text{SPCC} & \text{MSE} & \text{\emph{score}} & \text{SPCC}\\ 
    \midrule   
    (4-4-1) NN & 0.0 & 1.387e-02 & 0.9712 & 0.9862 & 1.422e-02 & 0.9705 & 0.9859 & 1.412e-02 & 0.9701 & 0.9857\\
    with ReLU & 0.05 & 2.527e-02 & 0.8808 & 0.9388 & 2.595e-02 & 0.8778 & 0.9375 & 2.602e-02 & 0.8769 & 0.9371\\
    25 coeffs. & 0.1 & 5.225e-02 & 0.6844 & 0.8282 & 5.430e-02 & 0.6737 & 0.8225 & 5.462e-02 & 0.6714 & 0.8213\\
    \midrule
    (4-4-1) NN & 0.0 & 3.893e-03 & 0.9882 & 0.9941 & 4.038e-03 & 0.9878 & 0.9939 & 3.985e-03 & 0.9878 & 0.9939\\
	with Tanh  & 0.05 & 2.143e-02 & 0.8960 & 0.9466 & 2.204e-02 & 0.8933 & 0.9455 & 2.212e-02 & 0.8929 & 0.9453\\
    25 coeffs. & 0.1 & 4.859e-02 & 0.6931 & 0.8325 & 5.024e-02 & 0.6863 & 0.8294 & 5.009e-02 & 0.6841 & 0.8282\\
    \midrule
    4-TT & 0.0 & 3.966e-03 & 0.9851 & 0.9926 & 3.990e-03 & 0.9851 & 0.9925 & 4.218e-03 & 0.9830 & 0.9915\\
    for (S=2, R=2) & 0.05 & 2.327e-02 & 0.8826 & 0.9395 & 2.722e-02 & 0.8611 & 0.9290 & 2.599e-02 & 0.8592 & 0.9271\\
    24 coeffs. & 0.1 & 5.174e-02 & 0.6743 & 0.8221 & 5.502e-02 & 0.6464 & 0.8042 & 5.403e-02 & 0.6380 & 0.7988\\
    \midrule
    (4-6-1) NN & 0.0 & 8.732e-03 & 0.9831 & 0.9921 & 8.980e-03 & 0.9826 & 0.9919 & 8.867e-03 & 0.9824 & 0.9918\\ 
    with ReLU & 0.05 & 2.207e-02 & 0.8946 & 0.9460 & 2.309e-02 & 0.8912 & 0.9446 & 2.275e-02 & 0.8903 & 0.9441\\
    37 coeffs. & 0.1 & 4.788e-02 & 0.6998 & 0.8369 & 5.021e-02 & 0.6900 & 0.8319 & 4.998e-02 & 0.6872 & 0.8305\\
    \midrule
    (4-6-1) NN & 0.0 & 4.206e-03 & 0.9893 & 0.9946 & 4.328e-03 & 0.9889 & 0.9945 & 4.367e-03 & 0.9888 & 0.9944\\
    with Tanh & 0.05 & 2.101e-02 & 0.8974 & 0.9473 & 2.131e-02 & 0.8954 & 0.9465 & 2.162e-02 & 0.8945 & 0.9461\\
    37 coeffs. & 0.1 & 4.877e-02 & 0.6963 & 0.8344 & 5.055e-02 & 0.6857 & 0.8290 & 5.072e-02 & 0.6851 & 0.8286\\
    \midrule
    4-TT & 0.0 & 2.677e-03 & 0.9900 & 0.9950 & 2.719e-03 & 0.9898 & 0.9949 & 2.962e-03 & 0.9881 & 0.9940\\
    for (S=2, R=4) & 0.05 & 2.140e-02 & 0.8920 & 0.9445 & 2.347e-02 & 0.8802 & 0.9387 & 2.300e-02 & 0.8752 & 0.9356\\
    40 coeffs. & 0.1 & 4.939e-02 & 0.6890 & 0.8301 & 5.209e-02 & 0.6651 & 0.8169 & 5.138e-02 & 0.6549 & 0.8096\\
    \midrule
    (4-15-1) NN & 0.0 & 3.306e-03 & 0.9918 & 0.9960 & 3.460e-03 & 0.9913 & 0.9958 & 3.437e-03 & 0.9913 & 0.9958\\    
    with ReLU & \color{red}0.05 & \color{red}1.915e-02 & \color{red}0.9039 & \color{red}0.9508 & \color{red}2.027e-02 & \color{red}0.8987 & \color{red}0.9483 & \color{red}2.035e-02 & \color{red}0.8975 & \color{red}0.9477\\
    91 coeffs. & \color{red}0.1 & \color{red}4.565e-02 & \color{red}0.7098 & \color{red}0.8425 & \color{red}4.818e-02 & \color{red}0.6933 & \color{red}0.8336 & \color{red}4.856e-02 & \color{red}0.6916 & \color{red}0.8326\\
    \midrule
    (4-15-1) NN & 0.0 & 3.880e-03 & 0.9904 & 0.9952 & 3.959e-03 & 0.9899 & 0.9950 & 3.945e-03 & 0.9901 & 0.9951\\    
    with Tanh & 0.05 & 2.118e-02 & 0.8997 & 0.9486 & 2.175e-02 & 0.8976 & 0.9478 & 2.174e-02 & 0.8963 & 0.9470\\
    91 coeffs. & 0.1 & 5.003e-02 & 0.6893 & 0.8302 & 5.178e-02 & 0.6820 & 0.8267 & 5.169e-02 & 0.6795 & 0.8255\\
    \midrule
    4-TT & \color{red}0.0 & \color{red}5.966e-04 & \color{red}0.9978 & \color{red}0.9989 & \color{red}5.857e-04 & \color{red}0.9978 & \color{red}0.9989 & \color{red}6.985e-04 & \color{red}0.9972 & \color{red}0.9986\\
    for (S=3, R=4) & 0.05 & 1.834e-02 & 0.9074 & 0.9526 & 1.888e-02 & 0.9036 & 0.9506 & 1.977e-02 & 0.8927 & 0.9450\\
    90 coeffs. & 0.1 & 4.567e-02 & 0.7125 & 0.8442 & 4.568e-02 & 0.7060 & 0.8404 & 4.708e-02 & 0.6842 & 0.8274\\
    \bottomrule
\end{tabular}}}
\end{table*}

\begin{table*}[!h]
\centering
\caption{Mackey-Glass time series for long-term prediction}\label{tab:mackeyglass_longterm}
{\scriptsize{\begin{tabular}{c c c c c c c c c c c}
    \toprule
    \multirow{2}{*}{\textbf{models}} & \multirow{2}{*}{$\sigma_{\mathrm{N}}$} & \multicolumn{3}{c}{\textbf{training}} & \multicolumn{3}{c}{\textbf{validation}} & \multicolumn{3}{c}{\textbf{test}}\\
    \cmidrule{3-11} 
    & & \text{MSE} & \text{\emph{score}} & \text{SPCC} & \text{MSE} & \text{\emph{score}} & \text{SPCC} & \text{MSE} & \text{\emph{score}} & \text{SPCC}\\ 
    \midrule
    (4-4-1) NN & 0.0 & 4.320e-02 & 0.8462 & 0.9209 & 4.561e-02 & 0.8382 & 0.9170 & 4.509e-02 & 0.8390 & 0.9174\\
    with ReLU & 0.05 & 4.502e-02 & 0.7789 & 0.8832 & 4.643e-02 & 0.7718 & 0.8800 & 4.610e-02 & 0.7736 & 0.8810\\
    25 coeffs. & 0.1 & 6.022e-02 & 0.6283 & 0.7934 & 6.192e-02 & 0.6231 & 0.7914 & 6.235e-02 & 0.6210 & 0.7899\\ 
    \midrule
    (4-4-1) NN & 0.0 & 4.009e-02 & 0.8468 & 0.9203 & 4.039e-02 & 0.8469 & 0.9208 & 4.098e-02 & 0.8437 & 0.9192\\
    with Tanh & 0.05 & 4.280e-02 & 0.7796 & 0.8830 & 4.375e-02 & 0.7770 & 0.8822 & 4.376e-02 & 0.7753 & 0.8813\\
    25 coeffs. & 0.1 & 5.823e-02 & 0.6286 & 0.7935 & 5.938e-02 & 0.6222 & 0.7902 & 5.942e-02 & 0.6211 & 0.7901\\
    \midrule
    4-TT & 0.0 & 3.456e-02 & 0.8679 & 0.9316 & 3.930e-02 & 0.8483 & 0.9211 & 4.691e-02 & 0.8218 & 0.9065\\      
   (S=2, R=2) & 0.05 & 4.264e-02 & 0.7806 & 0.8835 & 4.207e-02 & 0.7803 & 0.8834 & 4.125e-02 & 0.7937 & 0.8929\\       
    24 coeffs. & 0.1 & 5.950e-02 & 0.6187 & 0.7866 & 5.513e-02 & 0.6419 & 0.8014 & 5.255e-02 & 0.6755 & 0.8287\\
    \midrule
    (4-6-1) NN & 0.0 & 3.982e-02 & 0.8506 & 0.9223 & 4.174e-02 & 0.8443 & 0.9194 & 4.173e-02 & 0.8434 & 0.9190\\
    with ReLU & 0.05 & 4.342e-02 & 0.7822 & 0.8848 & 4.538e-02 & 0.7735 & 0.8805 & 4.500e-02 & 0.7749 & 0.8814\\
   37 coeffs. & 0.1 & 5.815e-02 & 0.6354 & 0.7973 & 6.055e-02 & 0.6225 & 0.7905 & 6.033e-02 & 0.6216 & 0.7899\\
    \midrule
    (4-6-1) NN & 0.0 & 3.9734e-02 & 0.8482 & 0.9211 & 4.110e-02 & 0.8425 & 0.9185 & 4.111e-02 & 0.8430 & 0.9187\\
    with Tanh & 0.05 & 4.297e-02 & 0.7800 & 0.8832 & 4.334e-02 & 0.7760 & 0.8816 & 4.357e-02 & 0.7768 & 0.8822\\
   37 coeffs. & 0.1 & 5.817e-02 & 0.6283 & 0.7929 & 5.870e-02 & 0.6274 & 0.7932 & 5.945e-02 & 0.6242 & 0.7914\\
    \midrule
    4-TT & 0.0 & 3.423e-02 & 0.8692 & 0.9323 & 3.891e-02 & 0.8498 & 0.9219 & 4.670e-02 & 0.8226 & 0.9070\\
   for (S=2, R=4) & 0.05 & 4.207e-02 & 0.7835 & 0.8852 & 4.113e-02 & 0.7853 & 0.8862 & 4.134e-02 & 0.7933 & 0.8926\\
   40 coeffs. & 0.1 & 5.902e-02 & 0.6218 & 0.7886 & 5.455e-02 & 0.6457 & 0.8039 & 5.256e-02 & 0.6755 & 0.8285\\      
    \midrule
    (4-15-1) NN & 0.0 & 3.782e-02 & 0.8566 & 0.9255 & 3.987e-02 & 0.8483 & 0.9215 & 3.985e-02 & 0.8489 & 0.9219\\
    with ReLU & 0.05 & 4.178e-02 & 0.7871 & 0.8872 & 4.389e-02 & 0.7763 & 0.8818 & 4.416e-02 & 0.7765 & 0.8819\\
    91 coeffs. & 0.1 & 5.642e-02 & 0.6400 & 0.8001 & 5.860e-02 & 0.6291 & 0.7943 & 5.916e-02 & 0.6269 & 0.7931\\
    \midrule
    (4-15-1) NN & 0.0 & 3.964e-02 & 0.8485 & 0.9212 & 4.064e-02 & 0.8451 & 0.9197 & 4.071e-02 & 0.8442 & 0.9194\\
    with Tanh & 0.05 & 4.287e-02 & 0.7800 & 0.8832 & 4.389e-02 & 0.7751 & 0.8810 & 4.355e-02 & 0.7770 & 0.8823\\
    91 coeffs. & 0.1 & 5.821e-02 & 0.6287 & 0.7931 & 5.928e-02 & 0.6230 & 0.7905 & 5.942e-02 & 0.6211 & 0.7895\\
    \midrule
    4-TT & \color{red}0.0 & \color{red}2.471e-02 & \color{red}0.9055 & \color{red}0.9518 & \color{red}2.797e-02 & \color{red}0.8920 & \color{red}0.9445 & \color{red}3.320e-02 & \color{red}0.8739 & \color{red}0.9355\\
   for (S=3, R=4) & \color{red}0.05 & \color{red}3.827e-02 & \color{red}0.8031 & \color{red}0.8962 & \color{red}3.733e-02 & \color{red}0.8064 & \color{red}0.8980 & \color{red}3.730e-02 & \color{red}0.8136 & \color{red}0.9040\\      
    90 coeffs. & \color{red}0.1 & \color{red}5.706e-02 & \color{red}0.6343 & \color{red}0.7966 & \color{red}5.367e-02 & \color{red}0.6532 & \color{red}0.8087 & \color{red}5.072e-02 & \color{red}0.6868 & \color{red}0.8374\\        
    \bottomrule
\end{tabular}}}
\end{table*}


\begin{landscape}
\begin{table*}
\centering
\caption{NASDAQ index for short-term prediction}\label{tab:nasdaq_shortterm}
{\scriptsize{\begin{tabular}{c c c c c c c c c c c c c c}
    \toprule
    \multirow{2}{*}{\textbf{models}} & {\textbf{no. of}} & \multicolumn{4}{c}{\textbf{training}} & \multicolumn{4}{c}{\textbf{validation}} & \multicolumn{4}{c}{\textbf{test}}\\
    \cmidrule{3-14} 
    & {\textbf{coeffs.}} & \text{MSE} & \text{\emph{score}} & \text{SPCC} & \text{\emph{$R$-squared}} & \text{MSE} & \text{\emph{score}} & \text{SPCC} & \text{\emph{$R$-squared}} & \text{MSE} & \text{\emph{score}} & \text{SPCC} & \text{\emph{$R$-squared}}\\
    \midrule
    \color{red}(4-4-1) NN ReLU & \color{red}25 & \color{red}8.948e-03 & \color{red}0.9395 & \color{red}0.9693 & \color{red}0.9395 & \color{red}1.130e-02 & \color{red}0.9222 & \color{red}0.9627 & \color{red}0.9200 & \color{red}1.134e-02 & \color{red}0.9212 & \color{red}0.9616 & \color{red}0.9191\\
    (4-15-1) NN ReLU & 91 & 7.931e-03 & 0.9464 & 0.9728 & 0.9464 & 1.203e-02 & 0.9174 & 0.9605 & 0.9150 & 1.234e-02 & 0.9147 & 0.9591 & 0.9124\\
    (4-30-1) NN ReLU & 181 & 7.394e-03 & 0.9500 & 0.9747 & 0.9500 & 1.220e-02 & 0.9163 & 0.9603 & 0.9138 & 1.256e-02 & 0.9132 & 0.9582 & 0.9108\\
    (4-91-1) NN ReLU & 547 & 6.662e-03 & 0.9550 & 0.9772 & 0.9549 & 1.285e-02 & 0.9118 & 0.9583 & 0.9093 & 1.328e-02 & 0.9086 & 0.9562 & 0.9062\\
    (4-217-1) NN ReLU & 1303 & 5.956e-03 & 0.9597 & 0.9796 & 0.9597 & 1.361e-02 & 0.9068 & 0.9561 & 0.9040 & 1.383e-02 & 0.9045 & 0.9543 & 0.9020\\
    \midrule
    \color{red}(4-4-1) NN Tanh & \color{red}25 & \color{red}9.295e-03 & \color{red}0.9371 & \color{red}0.9680 & \color{red}0.9371 & \color{red}1.059e-02 & \color{red}0.9268 & \color{red}0.9645 & \color{red}0.9246 & \color{red}1.087e-02 & \color{red}0.9243 & \color{red}0.9630 & \color{red}0.9222\\
    (4-15-1) NN Tanh & 91 & 9.181e-03 & 0.9379 & 0.9684 & 0.9378 & 1.066e-02 & 0.9263 & 0.9644 & 0.9240 & 1.120e-02 & 0.9223 & 0.9624 & 0.9201\\
    (4-30-1) NN Tanh & 181 & 9.164e-03 & 0.9380 & 0.9685 & 0.9380 & 1.064e-02 & 0.9264 & 0.9644 & 0.9242 & 1.124e-02 & 0.9220 & 0.9623 & 0.9199\\
    (4-91-1) NN Tanh & 547 & 9.159e-03 & 0.9380 & 0.9685 & 0.9380 & 1.068e-02 & 0.9260 & 0.9643 & 0.9238 & 1.123e-02 & 0.9221 & 0.9624 & 0.9200\\
    (4-217-1) NN Tanh & 1303 & 9.200e-03 & 0.9378 & 0.9684 & 0.9377 & 1.073e-02 & 0.9258 & 0.9642 & 0.9236 & 1.111e-02 & 0.9229 & 0.9627 & 0.9207\\
    \midrule  
    \color{red}4-TT (S=2,R=2) & \color{red}24 & \color{red}1.006e-02 & \color{red}0.9321 & \color{red}0.9656 & \color{red}0.9321 & \color{red}1.049e-02 & \color{red}0.9282 & \color{red}0.9647 & \color{red}0.9258 & \color{red}1.153e-02 & \color{red}0.9200 & \color{red}0.9610 & \color{red}0.9178\\      
    4-TT (S=3,R=4) & 90 & 1.009e-02 & 0.9319 & 0.9657 & 0.9319 & 1.100e-02 & 0.9246 & 0.9628 & 0.9221 & 1.307e-02 & 0.9099 & 0.9571 & 0.9073\\           
    4-TT (S=3,R=9) & 180 & 1.008e-02 & 0.9319 & 0.9657 & 0.9319 & 1.102e-02 & 0.9245 & 0.9627 & 0.9220 & 1.302e-02 & 0.9102 & 0.9572 & 0.9076\\        
    4-TT (S=4,R=16) & 544 & 1.020e-02 & 0.9310 & 0.9654 & 0.9310 & 1.137e-02 & 0.9221 & 0.9614 & 0.9195 & 1.383e-02 & 0.9050 & 0.9552 & 0.9022\\                           
    4-TT (S=5,R=25) & 1300 & 9.975e-03 & 0.9324 & 0.9661 & 0.9324 & 1.151e-02 & 0.9210 & 0.9612 & 0.9186 & 1.400e-02 & 0.9042 & 0.9550 & 0.9014\\ 
    \bottomrule
\end{tabular}}}
\end{table*}
 
\begin{table*}
\centering
\caption{NASDAQ index for long-term prediction}\label{tab:nasdaq_longterm}
{\scriptsize{\begin{tabular}{c c c c c c c c c c c c c c}
    \toprule
    \multirow{2}{*}{\textbf{models}} & {\textbf{no. of}} & \multicolumn{4}{c}{\textbf{training}} & \multicolumn{4}{c}{\textbf{validation}} & \multicolumn{4}{c}{\textbf{test}}\\
    \cmidrule{3-14} 
    & {\textbf{coeffs.}} & \text{MSE} & \text{\emph{score}} & \text{SPCC} & \text{\emph{$R$-squared}} & \text{MSE} & \text{\emph{score}} & \text{SPCC} & \text{\emph{$R$-squared}} & \text{MSE} & \text{\emph{score}} & \text{SPCC} & \text{\emph{$R$-squared}}\\ 
    \midrule
    (4-4-1) NN ReLU & 25 & 3.335e-02 & 0.8360 & 0.9115 & 0.8360 & 4.316e-02 & 0.7911 & 0.8976 & 0.7780 & 4.489e-02 & 0.7856 & 0.8933 & 0.7709\\
    (4-15-1) NN ReLU & 91 & 1.715e-02 & 0.9163 & 0.9572 & 0.9163 & 3.250e-02 & 0.8449 & 0.9289 & 0.8344 & 3.263e-02 & 0.8442 & 0.9279 & 0.8364\\
    \color{red}(4-30-1) NN ReLU & \color{red}181 & \color{red}1.358e-02 & \color{red}0.9336 & \color{red}0.9662 & \color{red}0.9336 & \color{red}3.169e-02 & \color{red}0.8503 & \color{red}0.9312 & \color{red}0.8401 & \color{red}3.158e-02 & \color{red}0.8501 & \color{red}0.9318 & \color{red}0.8421\\
    (4-91-1) NN ReLU & 547 & 9.291e-03 & 0.9548 & 0.9772 & 0.9548 & 3.315e-02 & 0.8417 & 0.9284 & 0.8315 & 3.231e-02 & 0.8469 & 0.9313 & 0.8386\\
    (4-217-1) NN ReLU & 1303 & 6.791e-03 & 0.9669 & 0.9833 & 0.9668 & 3.423e-02 & 0.8350 & 0.9260 & 0.8248 & 3.359e-02 & 0.8393 & 0.9283 & 0.8302\\
    \midrule
    (4-4-1) NN Tanh & 25 & 3.226e-02 & 0.8438 & 0.9178 & 0.8438 & 4.186e-02 & 0.7909 & 0.8987 & 0.7776 & 4.164e-02 & 0.8021 & 0.9038 & 0.7893\\
    (4-15-1) NN Tanh & 91 & 2.226e-02 & 0.8926 & 0.9443 & 0.8926 & 3.669e-02 & 0.8193 & 0.9154 & 0.8080 & 3.584e-02 & 0.8290 & 0.9202 & 0.8177\\
    (4-30-1) NN Tanh & 181 & 2.083e-02 & 0.8988 & 0.9477 & 0.8988 & 3.560e-02 & 0.8280 & 0.9206 & 0.8172 & 3.401e-02 & 0.8372 & 0.9248 & 0.8270\\
    (4-91-1) NN Tanh & 547 & 1.873e-02 & 0.9086 & 0.9532 & 0.9086 & 3.383e-02 & 0.8395 & 0.9268 & 0.8289 & 3.261e-02 & 0.8451 & 0.9303 & 0.8359\\
    \color{red}(4-217-1) NN Tanh & \color{red}1303 & \color{red}1.916e-02 & \color{red}0.9064 & \color{red}0.9520 & \color{red}0.9064 & \color{red}3.332e-02 & \color{red}0.8425 & \color{red}0.9282 & \color{red}0.8317 & \color{red}3.232e-02 & \color{red}0.8465 & \color{red}0.9308 & \color{red}0.8375\\
    \midrule      
    4-TT (S=2,R=2) & 24 & 3.689e-02 & 0.8194 & 0.9064 & 0.8192 & 4.282e-02 & 0.7952 & 0.9007 & 0.7814 & 4.453e-02 & 0.7883 & 0.8987 & 0.7767\\            
    4-TT (S=3,R=4) & 90 & 2.384e-02 & 0.8837 & 0.9409 & 0.8836 & 3.463e-02 & 0.8363 & 0.9226 & 0.8251 & 3.530e-02 & 0.8356 & 0.9227 & 0.8256\\     
    4-TT (S=3,R=9) & 180 & 2.345e-02 & 0.8853 & 0.9416 & 0.8852 & 3.479e-02 & 0.8354 & 0.9225 & 0.8242 & 3.514e-02 & 0.8361 & 0.9231 & 0.8262\\ 
    4-TT (S=4,R=16) & 544 & 2.078e-02 & 0.8986 & 0.9485 & 0.8986 & 3.267e-02 & 0.8466 & 0.9277 & 0.8361 & 3.357e-02 & 0.8438 & 0.9273 & 0.8341\\
    \color{red}4-TT (S=5,R=25) & \color{red}1303 & \color{red}1.944e-02 & \color{red}0.9053 & \color{red}0.9522 & \color{red}0.9052 & \color{red}3.196e-02 & \color{red}0.8503 & \color{red}0.9295 & \color{red}0.8398 & \color{red}3.319e-02 & \color{red}0.8458 & \color{red}0.9286 & \color{red}0.8355\\        
    \bottomrule
\end{tabular}}}
\end{table*}
\end{landscape}

\section*{Acknowledgements}
This work was supported by FAPESP, Brazil [grant number 2014/23936-4] and by CNPq, Brazil [grant number 308811/2019-4].

\bibliographystyle{elsarticle-num}
\bibliography{myrefs}

\end{document}